# Multi-Graph Convolutional-Recurrent Neural Network (MGC-RNN) for Short-Term Forecasting of Transit Passenger Flow

Yuxin He, Lishuai Li, Xinting Zhu, and Kwok Leung Tsui

*Abstract*—Short-term forecasting of passenger flow is critical for transit management and crowd regulation. Spatial dependencies, temporal dependencies, inter-station correlations driven by other latent factors, and exogenous factors bring challenges to the short-term forecasts of passenger flow of urban rail transit networks. An innovative deep learning approach, Multi-Graph Convolutional-Recurrent Neural Network (MGC-RNN) is proposed to forecast passenger flow in urban rail transit systems to incorporate these complex factors. We propose to use multiple graphs to encode the spatial and other heterogenous inter-station correlations. The temporal dynamics of the inter-station correlations are also modeled via the proposed multi-graph convolutional-recurrent neural network structure. Inflow and outflow of all stations can be collectively predicted with multiple time steps ahead via a sequence to sequence(seq2seq) architecture. The proposed method is applied to the short-term forecasts of passenger flow in Shenzhen Metro, China. The experimental results show that MGC-RNN outperforms the benchmark algorithms in terms of forecasting accuracy. Besides, it is found that the inter-station driven by network distance, network structure, and recent flow patterns are significant factors for passenger flow forecasting. Moreover, the architecture of LSTM-encoder-decoder can capture the temporal dependencies well. In general, the proposed framework could provide multiple views of passenger flow dynamics for fine prediction and exhibit a possibility for multi-source heterogeneous data fusion in the spatiotemporal forecast tasks.

## I. Introduction

SHORT-TERM forecasting of passenger flow is defined as the forecast with the short period, usually over 5 minutes and less than 1 hour [1]. Short-term forecasting of passenger flow in urban rail transit is a vital component for transit management and crowd regulation. A fine short-term forecast of passenger flow can support *transit operators* in optimizing service schedules, enhancing station passenger crowd regulation planning, thus adapting the supply transport the more precisely to fit the passenger demand, and also being aware of an emergency (an influx of passengers) and implementing emergency preparedness plans in advance [2]. Also, an accurate forecast of passenger flow information can help *passengers* to know where an influx of passengers in an immediate future will be and adjust their travel paths, modes, and departure times rationally.

However, short-term forecasting of passenger flow is a challenging task as the dynamics of passenger flow can be affected by many complex aspects, including spatial dependencies, temporal dependencies, inter-station correlations driven by other latent factors, and exogenous factors.

**Spatial dependencies.** Station-level passenger flow in urban rail transit networks is dominated by the topological structure of the transit network. Taking the network shown in Figure 1 as an example, the traffic status at station *a* is more related to station *c* than station *b*, and likewise, the traffic status at station *d* is more related to station *e* than station *f*, because the spatial correlations are based on network-based distance

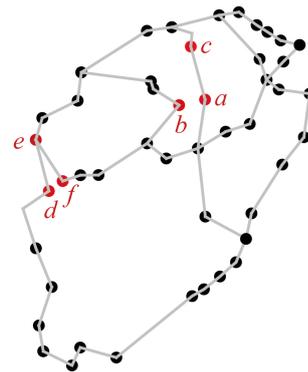

Fig. 1. Spatial dependencies of traffic flow in a transportation network.

Manuscript received ***; revised ***; accepted***. This work was supported by the Hong Kong Research Grants Council the General Research Fund (No. 11215119), National Social Science Foundation of China under Grant 20GBL301, and Guangdong Basic and Applied Basic Research Foundation under Grant 2021A1515110731.

Y. He is with the College of Urban Transportation and Logistics, Shenzhen Technology University, Shenzhen, China (e-mail: heyuxin@sztu.edu.cn).
L. Li is with the Section of Air Transport and Operations, Faculty of Aerospace Engineering, Delft University of Technology, and affiliated with School of Data Science, City University of Hong Kong, Hong Kong (e-mail: lishuai.li@tudelft.nl).
X. Zhu is with the School of Data Science, City University of Hong Kong, Hong Kong (e-mail: xtzhu3-c@my.cityu.edu.hk).
K. L. Tsui is with the Grado Department of Industrial and Systems Engineering, Virginia Polytechnic Institute and State University, Blacksburg, VA, USA (e-mail: kltsui@vt.edu).
Corresponding author: Lishuai Li

instead of Euclidean-based distance.

**Temporal dependencies.** Station-level passenger flow can be affected by different temporal features, including temporal autocorrelation, periodicity, and trend. For example, the traffic peak at 8 A.M. will affect that at 9 A.M. Besides, traffic conditions during morning rush hours present similar patterns on consecutive workdays. Moreover, traffic patterns may present a gradually increasing or decreasing trend due to seasonal reasons or some macro factors such as economics and policies.

**Inter-station correlations driven by other latent factors.** In addition to spatial correlations among stations dominated by the underlying network, stations can correlate with each other measured by multiple latent factors including static and dynamic factors. Static factors can be the network structural characteristics of stations (e.g., degree and betweenness centrality), operational information, and the functionality of stations. The dynamics of other accessible stations' passenger flow at previous time steps are a kind of dynamic factor. Taking the network in Figure 2 as an example, we can observe the dynamic inter-station correlations (orange arrow) in addition to temporal correlations (green arrow) and static inter-station correlations (blue arrow). The dynamic inter-station correlations indicate the influence across both the inter-station and temporal dimensions, between node $a$ and its accessible nodes at the next time step. Each node in the network can influence its accessible nodes at the same time step due to static inter-station correlations. Meanwhile, each node can also influence itself at the next time step due to the temporal correlations. Moreover, each node can even influence its accessible nodes at the next time step because of the dynamic inter-station correlations, as shown in Figure 2. The dynamic inter-station correlations result from the dynamical patterns such as the passenger flow evolving along both the stations and temporal dimensions simultaneously.

**Exogenous factors.** Exogenous factors, such as public holidays, day-of-week, weather, and big events may influence passenger flows. For example, passenger flow during the National Day Golden Week is much greater than that of regular days, and the distinguishment of day-of-week affects the passenger flow because of different trip purposes of different days (e.g., commuting and non-commuting travels). Moreover, extreme weather such as Typhoon may dramatically decrease passenger flow.

Some spatiotemporal deep learning approaches have been successfully applied to traffic forecasting in recent years, and some of them proposed the traffic forecasting methods based on a hybrid approach to extract spatiotemporal dependencies simultaneously. For example, the convolutional Long Short Term Memory (ConvLSTM) proposed by Shi, Chen, Wang, & Yeung (2015) [3] was used in spatiotemporal forecasting on transportation applications [4], [5]. The hybrid architectures show good performance on extracting spatiotemporal dependencies and correlations in forecasting, however, routine transportation activities commonly occur on a determined transportation network but not a Euclidean-based space [6], and Convolutional Neural Networks (CNNs) are commonly applied for dealing with Euclidean data [7] such as images, regular grids, and so on. Therefore, such hybrid models that combine CNN and Recurrent Neural Network (RNN) cannot describe the spatial features well when forecasting traffic flow in the context of a transportation network by taking the network topology into account. Graph Convolutional Networks (GCNs) were widely used to capture network-based spatial dependencies as GCNs can handle arbitrary graph-structured data. Several research works applied GCNs to capture network-based spatial dependencies in traffic forecasting tasks [8]–[10]. However, these studies only consider the topological relations between stations/points to build the graphs, but ignore all other latent factors that could measure the correlations (e.g., traffic patterns and local functionality). Chai et al. (2018) proposed a multi-graph convolutional neural network model to predict bike flow at station-level by considering heterogeneous inter-station relationships [11]. Lv et al. (2020) proposed a Temporal Multi-Graph Convolutional Network (T-MGCN) to jointly model the spatial, temporal, semantic correlations for traffic flow prediction [12]. However, they didn't consider the temporal dynamics of the inter-station correlations, specifically, the correlation regards traffic flow patterns are time-varying actually.

To fill in the aforementioned gaps, we propose an innovative deep learning approach, named Multi-Graph Convolutional-Recurrent Neural Network (MGC-RNN), to consider spatiotemporal dependencies and the complex inter-station correlations measured by static and dynamic factors simultaneously in the short-term forecasting of passenger flow. Specifically, we generate multiple graphs (including static and dynamic) to represent the inter-station correlations driven by different factors, respectively. Then we apply multiple GCNs to extract each graph's correlation information and then weighted-fuse all the extracted information. Our contributions are four-fold:

a) Multiple graphs encoding the spatial and heterogenous inter-station correlations are generated. In this way, besides temporal and spatial dependencies on the underlying transit networks, latent correlations driven by stations' features (e.g., similar traffic flow patterns, functionality, and network structure characteristics) are extracted. Besides, various correlation information is well represented by multi-dimensional adjacency matrices of multi-graphs.

b) The temporal dynamics of the inter-station correlations are considered since the inter-station correlations related to passenger flow patterns are time-varying. More specifically, the inter-station correlations may change along the temporal dimensions.

c) Inflow and outflow of all stations in an urban rail transit network can be collectively predicted with multiple time steps ahead via a sequence to sequence(seq2seq) architecture.

d) The proposed model is validated on a real-world case, Shenzhen metro passenger flow forecasting by incorporating spatiotemporal dependencies and multiple dimensional inter-station correlations. The proposed model structure is also flexible enough to handle several similar spatiotemporal forecast tasks.

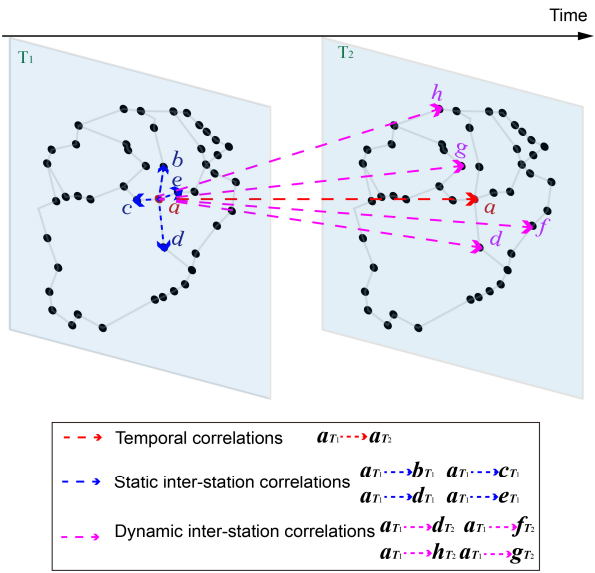

Fig. 2. The influence of node $a$ in the network. The green arrows indicate the influence of the node $a$ on itself at the next time step, i.e., the temporal correlations. The blue arrows indicate the influence of the node $a$ at its adjacent nodes $b, c, d, e$ at the current time step $T_1$, which represent the static inter-station correlations. The purple arrows indicate the influence across both the inter-station and temporal dimensions, between node $a$ and its accessible nodes $d, f, g, h$ at the next time step $T_2$.

The remainder of this paper is organized as follows. Section II reviews the prior research on traffic state forecasting and OD matrix forecasting. Section III provides the relevant notations and formulates the short-term OD matrix forecasting problem, and describes the methodology for short-term forecasting of passenger flow in urban rail transit. Section IV describes the real-world datasets used in the validation experiment and provides details for experiment settings and results analysis and discussion. Finally, Section V concludes this work.

## II. PRIOR RESEARCH

We summarized the related prior research on short-term traffic state forecasting in terms of the methods they used and what endogenous features they considered in Table I.

Table I shows that short-term traffic state forecasting consists of two categories according to the data features to be considered. The first category is formulated as time series prediction problems which only consider temporal dependencies of traffic state data. The second category is formulated as spatiotemporal data forecasting problems which consider both spatial and temporal dependencies.

For the above two categories of forecasting problems, a considerable number of studies were devoted to traffic state forecasting: and the existing methods can be categorized into two groups, including parametric methods (time series methods) and non-parametric methods. In time series analysis, Autoregressive Integrated Moving Average (ARIMA) and its family are the most general class of models for forecasting a time series. Hamed et al. (2002) applied ARIMA model to predict the short-term traffic volume on urban arterials [13]. Williams & Hoel (2003) presented the theoretical basis for modeling univariate traffic condition data streams as Seasonal Autoregressive Integrated Moving Average processes (SARIMA) [14].

Compared with time series methods (which are mainly parametric approaches), nonparametric models are flexible and sophisticated since their structure and parameters are not fixed. Higher prediction accuracy and more complex data modeling can be achieved by these models, such as k-Nearest Neighbors algorithm (KNN), Support Vector Machine (SVM), and Neural Networks (NNs). L. Zhang et al. (2013) presented a KNN model to predict short-term traffic flow [15]. Cai et al. (2016) proposed an improved KNN model to enhance forecasting accuracy based on spatiotemporal correlation and to achieve multi-step forecasting [16]. Besides, several researchers applied SVM to traffic prediction and also proposed improved-version SVM, e.g., chaos wavelet analysis SVMs [17], least-squares SVMs [18], particle swarm optimization SVMs [19], and genetic algorithm SVMs [20]. However, the biggest limitation of SVMs lies in the choice of the kernel. NNs and deep learning approaches have achieved a better performance in the domain of computer vision [21], [22]. Recurrent Neural Networks (RNNs) have made success for sequence learning tasks [23]. Involving long short-term memory (LSTM) or gated recurrent unit (GRU) enables RNNs to capture the long-term temporal dependency of time series. Ma et al. (2015) validated the effectiveness of LSTM by using Travel speed data in Beijing and demonstrated that LSTM could achieve the best prediction performance in terms of both accuracy and stability [24]. CNNs [25] especially the one-dimensional CNNs have been applied to sequences for decades [26]. Bai, Kolter, & Koltun (2018) proposed a generic temporal convolutional network (TCN) architecture that is applied to sequence modeling and found it demonstrated longer effective memory [27]. Some studies later proposed deep learning frameworks based on TCN in short-term traffic forecasting [28], [29].

However, these algorithms cannot capture the spatial features. Thanks to the powerful ability in capturing the correlations in the spatial domain, CNN is now widely used in learning from spatial data [25]. Especially for data types of spatial maps and spatial rasters, which can be represented as a two-dimensional matrix. Spatial dependencies are fundamental features to be extracted from traffic data, thus CNN is well suited to learn the spatial features of them [30-33]. When both spatial and temporal dependencies are considered in traffic forecasting tasks, traffic data are naturally spatiotemporal data, which are sometimes represented as a tensor or a sequence of tensors, three-dimensional CNNs [34] can be used to learn the complex spatial and temporal dependencies of the data.

Moreover, if the form of the input traffic data is a sequence of image-like matrices, hybrid models that combine CNN and RNN can be used to model the input. Some researchers proposed the forecasting methods based on a hybrid approach to extract spatiotemporal dependencies simultaneously in short-term traffic forecasting tasks. For example, Wu & Tan (2016) [7] and Yu, Wu, Wang, Wang, & Ma (2017) [35] proposed the structures with the combination of CNN and LSTM for spatiotemporal forecasting. Instead of simply stacking the architectures of CNN and RNN, by extending the fully connected LSTM (FC-LSTM) to have convolutional structures in both the input-to-state and state-to-state transitions, Shi, Chen, Wang, & Yeung (2015) proposed the convolutional LSTM (ConvLSTM) and used it to build an end-to-end trainable model for the precipitation nowcasting problem [3]. ConvLSTM was then used in spatiotemporal forecasting on transportation applications [4], [5]. The hybrid architectures show good performance in extracting spatiotemporal dependencies and correlations in forecasting, but the training procedure may become time-consuming as the size of the dataset increases as RNNs are affected by the size of sequences. Furthermore, J. Zhang et al. (2018) proposed a SpatioTemporal Residual Network (ST-ResNet) for forecasting crowd flow in each regular region of a city, yet it cannot be adapted to deal with the forecasts in irregular-shaped regions or Non-Euclidean space [36].

CNNs are commonly applied for dealing with Euclidean data such as images, regular grids, etc. However, spatial features based on the topological structure of a network or a graph have strong effects on modeling graph-structured data. Graph Convolutional Networks (GCNs) were widely used to capture network-based spatial dependencies as GCNs can handle arbitrary graph-structured data. In transportation fields, GCNs were popularly applied in traffic prediction tasks. Zhao, Song, Deng, & Li (2018) developed a spatiotemporal neural network named Temporal Graph Convolutional Network (TGCN) for forecasting problems, which combines the GCN with GRU [8]. Sun, Zhang, Li, Yi, & Zheng (2019) proposed a Multi-View Graph Convolutional Network to predict the inflow and outflow in each irregular region of a city [37]. Besides, another graph neural network, similar to GCNs, Diffusion Convolutional Neural Networks (DCNNs) were also developed for graph-structured data[38]. Later on, Li, Yu, Shahabi, & Liu (2017) proposed to model the traffic flow as a diffusion process on a directed graph and introduce Diffusion Convolutional Recurrent Neural Network (DCRNN), which is a deep learning framework for traffic forecasting that incorporates both spatial and temporal dependency in the traffic flow [39]. Besides, attention mechanism was also widely applied to the extraction of temporal and graph-structured spatial dependencies. Spatiotemporal Graph Attention Networks (GAT) was proposed to finish the tasks of spatiotemporal forecasts of traffic states [26], [40]. However, the abovementioned studies only considered the topological relations between stations/points of underlying transportation networks to build the graphs, but ignored all other latent factors that could measure the correlations (e.g., traffic patterns and local functionality).

To consider other latent factors influencing inter-station correlations, some related-studies came up with thoughts of using multi-graph convolutional neural networks to extract inter-station correlations measured by other factors. Chai et al. (2018) proposed a multi-graph convolutional neural network (M-GCN) model to predict bike flow at station-level by considering heterogeneous inter-station relationships [11]. Lv et al. (2020) proposed a Temporal Multi-Graph Convolutional

TABLE I
SUMMARY OF PRIOR RESEARCH ON SHORT-TERM TRAFFIC STATE FORECASTING

| Authors (Year) | Method | Endogenous features considered | | | | |
| --- | --- | --- | --- | --- | --- | --- |
| | | Temporal dependency | Spatial dependency | Euclidean-distance or Network-distance | Static latent factors | Dynamic latent factors |
| **Hamed et al. (2002)** | ARIMA | √ | - | - | - | - |
| **Williams & Hoel (2003)** | SARIMA | √ | - | - | - | - |
| **L. Zhang et al. (2013)** | KNN | √ | - | - | - | - |
| **Wang & Shi (2013); Cong et al. (2016); Gu et al. (2012); Yang et al. (2014)** | SVMs | √ | - | - | - | - |
| **Ma et al. (2015)** | LSTM | √ | - | - | - | - |
| **Ren et al. (2020); W. Zhao et al. (2019)** | TCN | √ | - | - | - | - |
| **Cai et al. (2016)** | KNN | √ | √ | E | - | - |
| **Wu & Tan (2016); Yu, Wu, Wang, Wang, & Ma (2017)** | Stack CNN and RNN | √ | √ | E | - | - |
| **Ai et al. (2019); Ke et al. (2017)** | ConvLSTM | √ | √ | E | - | - |
| **J. Zhang et al. (2018)** | ST-ResNet | √ | √ | E | - | - |
| **Zhao, Song, Deng, & Li (2018)** | TGCN | √ | √ | G | - | - |
| **Li, Yu, Shahabi, & Liu (2017)** | DCRNN | √ | √ | G | - | - |
| **Park et al. (2020); Wei & Sheng (2020)** | GAT | √ | √ | G | - | - |
| **Chai et al. (2018)** | M-GCN | √ | √ | G | √ | - |
| **Lv et al. (2020)** | T-MGCN | √ | √ | G | √ | - |
| **Proposed model** | MGC-RNN | √ | √ | G | √ | √ |

Network (T-MGCN) to jointly model the spatial, temporal, semantic correlations for traffic flow prediction [12]. However, they didn't consider the temporal dynamics of the inter-station correlations, specifically, they treated the correlation regards traffic flow patterns as static. Instead, the inter-station correlations driven by traffic patterns are time-varying actually because of the temporal dynamics of traffic patterns.

### III. METHODOLOGY

This paper aims to forecast inflow and outflow collectively of a certain station in an urban rail transit network in the immediate future based on the historical passenger flow information and available factors including the information of network distance, network structure, operational information, and stations' local functionality.

#### A. Notations and Problem Statement

In this subsection, we provide notations and he problem statement of this study. For easy retrieval, we summarize the notations used in this study in Table II.

TABLE II
NOTATIONS USED IN THIS STUDY

| Notations | Descriptions |
|---|---|
| $G$ | A graph |
| $V$ | The set of nodes in a graph |
| $E$ | The set of edges in a graph |
| $N$ | The number of nodes |
| $M \in \mathbb{R}^{N \times N}$ | The adjacency matrix |
| $G^1$ | The network distance graph |
| $G^2$ | The Point-Of-Interest (POI) correlation graph |
| $G^3$ | The network structure correlation graph |
| $G^4$ | The operational information correlation graph |
| $G_t^5$ | The recent flow correlation graph at time $t$ |
| $M^1 \in \mathbb{R}^{N \times N}$ | The network distance adjacency matrix |
| $M^2 \in \mathbb{R}^{N \times N}$ | The POI correlation adjacency matrix |
| $M^3 \in \mathbb{R}^{N \times N}$ | The network structure correlation adjacency matrix |
| $M^4 \in \mathbb{R}^{N \times N}$ | The operational information correlation adjacency matrix |
| $M_t^5 \in \mathbb{R}^{N \times N}$ | The recent flow correlation adjacency matrix at time $t$ |
| $S$ | The passenger flow type of each station (i.e., $S = 2$, referring to inflow and outflow, respectively) |
| $Y_t \in \mathbb{R}^{N \times S}$ | The passenger flow of each station of the network at time $t$ |
| $d_t$ | The day-of-week |
| $h_t$ | The holiday-or-not |
| $o_t$ | The hour-of-day |
| $A_t$ | The weather information |

**Multiple graphs generated from different aspects.** A transit network graph is denoted as $G(V, E, M)$, where $V = \{v_1, v_2, \ldots, v_N\}$ represents a set of nodes representing stations in the transit network, and $N$ refers to the number of stations. $E \ni e_{ij}$ denotes a set of edges from node $i$ to $j$, $i, j \in V$, indicating the correlation between node $i$ and $j$. $M \in \mathbb{R}^{N \times N}$ denotes the adjacency matrix, indicating the weights of two correlated nodes. The larger the weight, the higher the two stations correlate with each other. The proposed model is able to incorporate multiple graphs according to the different perspectives of correlations among stations. In this study, we include network distance graph $G^1$, Point-Of-Interest (POI) correlation graph $G^2$, network structure correlation graph $G^3$, operational information correlation graph $G^4$, and recent flow correlation graph $G_t^5$ at time $t$. Corresponding to each of the graphs, we denote the network distance adjacency matrix, POI correlation adjacency matrix, network structure correlation adjacency matrix, and operational information correlation adjacency matrix as $M^1, M^2, M^3, M^4 \in \mathbb{R}^{N \times N}$. Since we consider the dynamic inter-station correlations, the recent flow correlation adjacency matrix is time-varying, and the recent flow correlation adjacency matrix at time $t$ is denoted as $M_t^5 \in \mathbb{R}^{N \times N}$.

**Historical inflow & outflow sequence.** The passenger flow on the transit network changes dynamically over time. It is denoted as $Y_t \in \mathbb{R}^{N \times S}$ of each station of the network at time $t$, where $S$ denotes the passenger flow type of each station (i.e., $S = 2$, referring to inflow and outflow, respectively). Given input sequence length $l$, the historical inflow and outflow sequence can be denoted as $[Y_{T-l}, Y_{T-l+1}, \ldots, Y_{T-1}]$.

**Exogenous factors.**

a) Day-of-week. We use a categorical variable denoted by $d_t$ to represent the day-of-week. $d_t$ captures the distinguished properties between each day of the week. If $t$ belongs to the $r$th day of the week, then
$$d_t = r, \quad r = 1,2,\ldots,7 \qquad (1)$$

b) Holiday. We also denote another dummy variable $h_t$ to be the holiday-or-not, which distinguishes whether it is a holiday (includes adjacent weekends) or not.
$$h_t = \begin{cases} 1, & \text{if } t \text{ belongs to holidays} \\ 0, & \text{otherwise} \end{cases} \qquad (2)$$

c) Hour-of-day. We use another categorical variable denoted by $o_t$ to represent the hour-of-dat. $o_t$ captures the distinguished properties between each hour of the day. Note that we only consider the opening hours of subway in one day, e.g., the opening hours of Shenzhen metro system in 2013 of all 5 lines are 6:30-23:00, so there are 17 hourly period during the opening hours per day. If $t$ belongs to the $s$th hour of the day, then
$$o_t = s, \quad s = 1,2,\ldots,17 \qquad (3)$$

d) Weather. The weather variable has 3 dimensions including the weather conditions with 8 types, temperature, and wind speed during the $t$-th time interval, which is denoted as $A_t = (a_t^1, a_t^2, a_t^3)$. The first dimension $a_t^1$ is a categorical vector distinguishing eight weather conditions clear, sunny, fog, cloudy, light rain, rain shower, thunderstorm, and overcast, and the last two digits $(a_t^2, a_t^3)$ are numeric vectors that denote temperature and wind speed respectively.

**Problem Statement.**

Given five generated graph $G^1(V, E, M^1)$, $G^2(V, E, M^2)$, $G^3(V, E, M^3)$, $G^4(V, E, M^4)$, and $G_t^5(V, E, M_t^5)$ and the historical observations and pre-known information $\{Y_t | t = 0, \ldots, T-1; d_t, h_t, o_t | t = 0, \ldots, T; A_t | t = 0, \ldots, T-1\}$, predict $[Y_T, Y_{T+1}, \ldots, Y_{T+p}]$. Note that the day-of-week $d_t$, holiday $h_t$, and hour-of-day $o_t$ of the $t$-th time interval are pre-known at $t$.

#### B. Overall Framework

Figure 3 shows the overall procedures of the proposed framework, which includes six parts: raw data, feature engineering, input layer, graph convolutional layer,

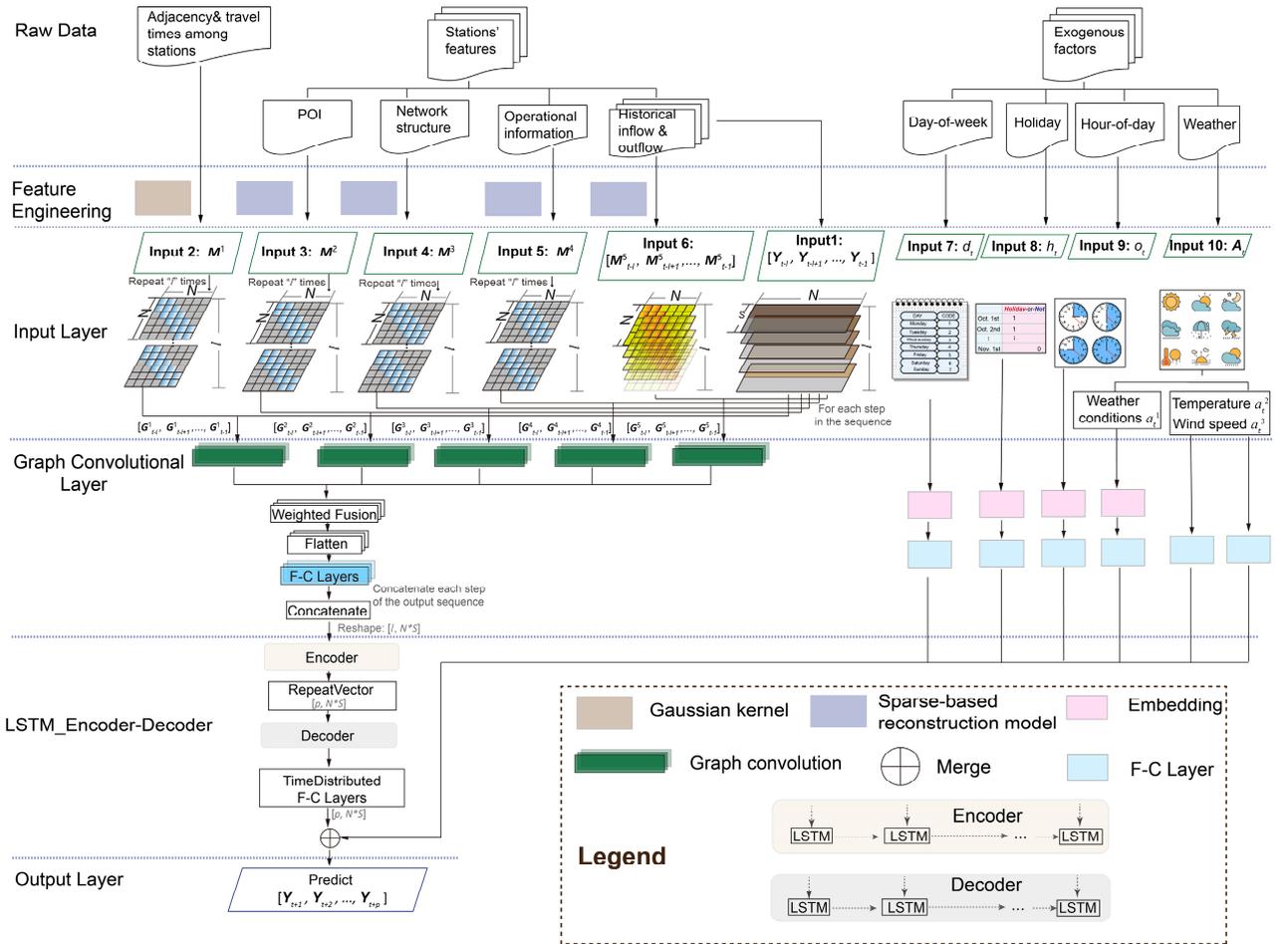

Fig. 3. The architecture of MGC-RNN.

LSTM_encoder-decoder, and output layer.

First of all, raw data including adjacency and travel times among stations, and stations' features including POI information, network structure, operational information, and historical inflow and outflow are fed into the framework, as shown in Figure 3. The raw data are then processed through feature engineering procedures, wherein, stations' features are used to calculate different adjacency matrices via a sparse-based reconstruction model [41], as shown in Input 3 to input 6, denoted by $M^2, M^3, M^4$, and $M^5$. Input 2, network distance adjacency matrix $M^1$ was generated with the information of adjacency and travel times among stations, here, the weights were calculated by a Gaussian kernel function. Input 1 is the historical inflow and outflow sequence, $[Y_{t-l}, Y_{t-l+1}, ..., Y_{t-1}]$, and the sequence length is $l$, where $Y_T$ has the dimension of $N \times S$. $N$ denotes the number of stations, $S$ equals 2 representing inflow and outflow, i.e., the two types of passenger flows. They were concatenated by using the raw historical passenger flow data. Since $M^1, M^2, M^3, M^4$ are static information, we need to repeat them $l$ times to make them as a sequence of equal length with other input sequences, i.e., input 1 and input 6. Input 6 is recent flow correlation adjacency matrix sequence $[M^5_{t-l}, M^5_{t-l+1}, ..., M^5_{t-1}]$, as mentioned earlier, the optimal reconstructed coefficient matrix $M^5$ is time-varying, so they naturally make up a sequence. The sequence of recent flow correlation adjacency matrix ensures the temporal dynamics of inter-station correlations to be considered.

The graph convolutional layer, for each time step in the sequences, we combine each adjacency matrix with each inflow and outflow at each time step as the topological structure and node's feature of a graph, and then conduct five graph convolution operators for all these input graphs, then use a self-defined weighted fusion layer to merge the outputs from the five graphs, and the weights here are learnable. In this way, we extract the network-based spatial dependencies among stations as well as static and dynamic inter-station correlations.

Then via flatten and fully-connected layers, we concatenate the output at each time step and reshape it as the input to the LSTM network, here the size of LSTM input is $[l, N \times S]$. We would like to conduct multi-step forecasting, so we apply one of the seq2seq structures, i.e., LSTM encoder-decoder structure. We can see encoder and decoder are all LSTM units. The encoder is to read and encodes the input sequence, and the decoder will read the encoded input sequence. After encoding by the encoder, the internal representation of the input sequence is repeated multiple times, once for each time step in the output sequence. This sequence will be presented to the LSTM decoder. To interpret each time step in the output sequence, the interpretation layer and the output layer are wrapped in a TimeDistributed wrapper and reuse the same weights to perform the interpretation.

There are four parallel structures for modeling the categorical exogenous factors (i.e., day-of-week, holiday information, hour-of-day, and weather conditions), where each structure has two layers, and the first layer serves as an embedding layer for keeping the size of the input vector much smaller than the huge one-hot encoded vector. The second layer is an F-C layer. As the wind speed and temperature are numeric vectors, we will use F-C layer to process them directly without embedding. The outputs of the exogenous factors unit are then merged (adding operation) with the output from LSTM_encoder-decoder structure.

Finally, we can see we output the $p$-step outputs, $Y_t$, $Y_{t+1}, \ldots, Y_{t+p}$.

### C. Multigraph Convolution

To fully encode the relationship between stations, we generate multiple graphs that contain heterogeneous spatial and inter-station correlations and employ multiple parallel graph convolutional operators on multigraphs in our neural network model.

*1) Multigraph Generation*

Graph generation is the fundamental step for GCNs. Stations in transit networks often correlate with each other in different aspects. The higher correlations between stations, the larger the weight to be assigned. On this basis, we innovatively generate five alternative graphs in terms of different aspects, including network distance graph, POI correlation graph, network structure correlation graph, operational information correlation graph, and recent flow correlation graph.

a) **Network distance graph $G^1$**. In practice, since the distance graph is generated to model the spatial correlations among stations, it is necessary to consider the network layout when calibrating the distance, however, existing studies usually just use Euclidean-based distance metric to calibrate the distance no matter how the locations connected. In practice, in subway networks, one station is connected to another station with the railway, so the Euclidean-based distance cannot reflect the actual distance between two stations. The study uses network-based distance to measure spatial connectivity instead of Euclidean-based distance [42]. Also, considering the actual travel costs that passengers take, we use the average of the travel times between two stations as the network-based distance. Specifically, we take the average of the travel times between two stations during different periods (per hour period) of workday and weekend from AMap[1] by checking the travel time of the routes by transit mode. After getting the network-based distance $d_{ij}$ between station $i$ and $j$, we calculate the weight $W_{ij}$ between $i$ and $j$ of the adjacency matrix by a Gaussian kernel function as (4).

i. $$W_{ij} = \exp(-\frac{d_{ij}^2}{\sigma^2}) \quad (4)$$

where $\sigma$ is the standard deviation of distances.

b) **POI correlation graph $G^2$**. When forecasting the passenger flow for a station, it is intuitive to other stations that perform similar functionality with this station. The functionality of the station can be measured by POI information around the station. The edges' weights of POI correlation graph are calculated via a sparse-based reconstruction model proposed by S. Zhang et al. (2017) [41].

c) **Network structure correlation graph $G^3$**. Stations could also correlate with each other in terms of the characteristics of network structure. Network structure includes degree and betweenness of stations serving as the nodes in metro networks, and days since the stations opened, as well as spatial characteristics of each station: distance to the city center. In the field of complex networks, as degree is a simple centrality measure that counts how many neighbors a node has, and the betweenness centrality for each node refers to the number of shortest paths that pass through the node [43], thus they are correlated to the information for transfer stations or terminal stations, and the importance of stations in the aspect of their controlling over flows passing between others of metro networks. Days since stations opened reflect not only the operation time of stations but also the line information of stations. The edges' weights of network structure correlation graph are also calculated via the sparse-based reconstruction model.

d) **Operational information correlation graph $G^4$**. Stations with similar operational patterns could also correlate with each other. Specifically, the information of metro line's headways of peak hours and off-peak hours is treated as each metro station's operational features. Note that for the transfer station, the smaller headway is taken as its headway. The edges' weights of operational information correlation graph are also calculated via the sparse-based reconstruction model.

e) **Recent flow correlation graph $G_t^5$**. Last but not the least, stations with similar traffic patterns could also correlate with each other. Specifically, the correlation between stations in terms of similar traffic patterns is measured by the historical passenger flow of the stations. Instead of taking all historical inflow and outflow [11], [12], we take the recent 10 time intervals' inflow and outflow $X_t \in \mathbb{R}^{20 \times N}$ as features of the current time $t$ to measure the recent flow correlation among stations. In this way, the passenger flow correlations captured are time-varying but not static, which accords with the actual conditions. The edges' weights of the recent flow correlation graph at each time slot are also calculated via the sparse-based reconstruction model.

*i. Sparse-based reconstruction model*

As mentioned earlier, we calculate the POI correlation adjacency matrix $M^2$, network structure correlation adjacency matrix $M^3$, operational information correlation adjacency matrix $M^4$, and recent flow correlation adjacency matrix $W_T^5$ at each time slot $T$ via a sparse-base reconstruction model proposed by S. Zhang et al. (2017).

The principle of the sparse-base reconstruction model is as follows: given the observed feature matrix $X \in \mathbb{R}^{P \times N} = [x_1, \ldots, x_N]$, where $P$ denotes the number of features of each node. The original attribute matrix $X$ is used to reconstruct each node's attribute vector $x_i$, with the objective that minimizing the distance between $\mathbf{XW}$ and $\mathbf{X}$, where $W = [w_1, ..., w_N] \in \mathbb{R}^{N \times N}$ represents the reconstructed coefficient matrix or the correlations between nodes' features and themselves, i.e., the adjacency matrix we are aiming to. The objective function for the reconstruction process is as (5):

$$\min_W \{\|\mathbf{XW} - \mathbf{X}\|_F^2 + \rho_1 \|\mathbf{W}\|_1 + \rho_2 \text{Tr}(\mathbf{W}^T\mathbf{X}^T\mathbf{LXW})\},$$
$$\mathbf{W} \geq 0 \quad (5)$$

---

[1] Source: https://www.amap.com/

Where $\rho_1$ and $\rho_2$ are tuning parameters, and $\mathbf{L} \in \mathbb{R}^{P \times P}$ is a Laplacian matrix, which indicates the relational information between features. As shown in (4), the first part $\|\mathbf{XW} - \mathbf{X}\|_F^2$ is the least square loss function to measure the distance between $\mathbf{XW}$ and $\mathbf{X}$. The second term $\rho_1 \|\mathbf{W}\|_1$ is an $\ell 1$-norm regularization term, which is used to generate the sparse reconstruction coefficient. The third term of (4) is also a regularization term with the assumption that, if some features are related in regressing, then their predictions are also related to each other. Thus, their corresponding predictions should have the same or similar relation. To utilize such relation, the relation among nodes in $\mathbf{X}$ to be reflected in the relationship between their predictions is imposed, and through the derivation (see details in [41]), the third term is obtained.

All of the three terms as well as the nonnegative constraint in (4) are convex, thus the final objective function is convex. On this basis, the optimal reconstructed coefficient matrix, i.e., the adjacency matrix can be obtained through solving convex optimization.

ii. Gaussian kernel function

The weight $W_{ij}^1$ between station $i$ and station $j$ of the network distance adjacency matrix $M^1$ is calculated from a kernel function. Gaussian kernel is a popular function to calculate weight matrix in spatial models [44]. The weight calculated from Gaussian kernel function continuously and gradually decreases from the center of the kernel. Gaussian kernel is expressed as (3). If the stations $i$ and $j$ satisfy $d_{ij} = 0$, $W_{ij} = 1$, whereas $W_{ij}$ decreases according to a Gaussian curve as $d_{ij}$ increases.

2) Graph Convolutional Network (GCN)

In the proposed MGC-RNN model, parallel GCNs are employed to extract spatial correlations and inter-station correlations in terms of multiple aspects of multiple graphs. In our study, we mainly focus on the spectral framework to apply convolutions in spectral domains, which is named the spectral graph convolution [45]. Figure 4 presents a comparison between standard convolution and graph convolution. To better compare, we can treat images as a graph, i.e., each pixel of images can be represented as a node of the graph. As depicted in Figure 4(a), the neighbor nodes of the orange node are determined by a 3*3 filter, which is applied to take the weighted average of the pixel values of the orange node and its neighbor nodes. The ordering of the orange node's neighbor nodes is based on their positions, and the weights of neighbor nodes can be shared over all the positions of nodes. Similarly, a spectral graph convolution is the multiplication of a graph signal with a filter in the Fourier space of a graph, i.e., it also takes the average of the central node's attributes and its neighbor nodes' attributes. As shown in Figure 4 (b), assuming that the orange node is the central node, GCN can obtain the topological relationship between the central node and its neighbor nodes, so that it can learn the spatial features reflected by a network.

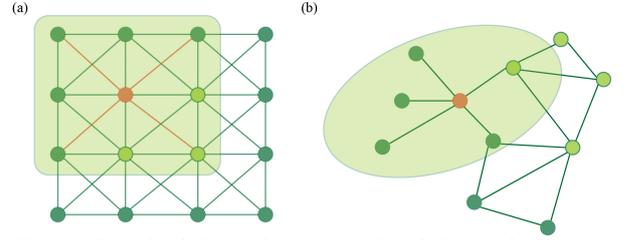

Fig. 4. Standard Convolution vs. Graph Convolution. (a) Standard Convolution. (b) Graph Convolution.

The idea of GCN is to realize convolution operation on the topological graph with the help of graph theory. Next, the core part of the GCN is introduced. Take a simple form of a layer-wise propagation rule as an example:

$$H^{(l+1)} = \sigma(AH^{(l)}W^{(l)}) \quad (6)$$

Wherein, $H^{(l)}$ denotes the $l$-th neural network layer with a weight matrix $W^{(l)}$. $A$ represents an adjacency matrix and $\sigma(\cdot)$ refers to a non-linear activation function such as $RELU$.

Since the node itself cannot be included when counting in all neighbor nodes' features by multiplication with $A$, so an identity matrix is added to $A$ to fix this limitation, and a matrix with a self-loop structure $\hat{A} = A + I$ is obtained, where $I$ is the identity matrix. Moreover, multiplication with $A$ which is not normalized will change the scale of the feature vectors, so a symmetric normalization is applied to address this issue, i.e., $\tilde{A} = \hat{D}^{-\frac{1}{2}}\hat{A}\hat{D}^{-\frac{1}{2}}$. Above all, the propagation rule can be expressed as:

$$H^{(l+1)} = \sigma\left(\hat{D}^{-\frac{1}{2}}\hat{A}\hat{D}^{-\frac{1}{2}}H^{(l)}W^{(l)}\right) \quad (7)$$

Where $\hat{D}$ denotes the diagonal node degree matrix of $\hat{A}$.

3) Fusion of the Outputs of Multigraph Convolution

Inter-station correlations can be influenced by multiple aspects (multiple graphs) and the degree can be different. Therefore, we apply a weighted fusion method as (8) in the model:

$$g_{merge} = W^1 \circ g^1 + W^2 \circ g^2 + W^3 \circ g^3 + W^4 \circ g^4 + W^5 \circ g^5$$
$$\in \mathbb{R}^{N \times U} \quad (8)$$

wherein, ° denotes Hadamard product (i.e. element-wise product), $W^1$, $W^2$, $W^3$, $W^4$, and $W^5$ are the parameters that can be learned, which quantify the degrees affected by the feature extracted from the five graphs above, denoted by $g^1$, $g^2$, $g^3$, $g^4$, and $g^5$ respectively. Note that, the weighted fusion was applied on each step of the output sequences, and the merged output is also one step of the sequence, which has the dimension of $N \times U$, where $U$ is the number of graph convolutional units.

D. LSTM_Encoder-Decoder

In this study, we conduct multi-step passenger flow forecasting because it can provide multiple time steps of predictions for operators and passengers, enabling them to have more time to react and take action. The Seq2Seq architecture is applied to tackle the multi-step forecasting in our study, which is shown in Figure 5.

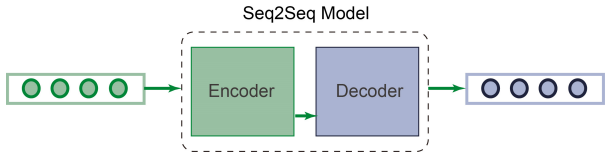

Fig. 5. The structure of Seq2Seq model.

The structure of seq2seq is comprised of two sub-models: one is the encoder that reads the input sequences and compresses them to a fixed-length internal representation, and an output model is the decoder that interprets the internal representation and uses it to predict the output sequence. We can use the network structure of LSTM or CNN as the components of the encoder and decoder. The strength of LSTM is in tackling sequence input, supporting multivariate inputs, and mapping input to output vector that may represent multiple output time steps. On this basis, we adopt the structure that LSTMs being both encoder and decoder.

LSTM_Encoder-Decoder is the structure whose encoder and decoder are all LSTM units [46]. As shown in Figure 3, the LSTM encoder is to read and encodes the input sequence as the internal representation, which is then repeated multiple times, once for each time step in the output sequence. The output sequence will be presented to the LSTM decoder.

### E. Loss

Mean Square Error(MSE) and Mean Absolute Error(MAE) are the most commonly used regression loss functions. MSE is greater for learning the outliers in the dataset, on the other hand, MAE is good to ignore the outliers. However, the shortcoming of using MAE as a loss function for the training of neural networks is its constantly large gradient, which can lead to missing minima at the end of training using gradient descent. Additionally, a model using MSE will make predictions skewed towards outliers. In some cases, the data which looks like outliers not bothered and also those points should not get high priority. Huber loss can be really helpful in such cases, as it curves around the minima which decreases the gradient, and it's more robust to outliers than MSE. Therefore, Huber loss is the combination of both MSE and MAE. The Huber loss is defined by:

$$\mathcal{L}_\delta(y, \hat{y}) = \begin{cases} \frac{1}{2}(y - \hat{y})^2, & \text{for } |y - \hat{y}| \leq \delta, \\ \delta|y - \hat{y}| - \frac{1}{2}\delta^2, & \text{otherwise.} \end{cases} \quad (9)$$

where $\delta$ is the hyperparameter to define the range for MAE and MSE. Huber loss approaches MSE when $\delta \sim 0$ and MAE when $\delta \sim \infty$ (large numbers.). We manually tuned the $\delta$ value here.

In this way, MGC-RNN can be trained with the following optimization objective:

$$\arg\min_{\Theta} \sum_{t \in T} \sum_{i=1}^{N} \sum_{s=1}^{S} \mathcal{L}_\delta(Y_t[i,s], \hat{Y}_t[i,s]) \quad (10)$$

Where $\Theta$ denotes all the learnable parameters of MGC-RNN. $Y_t[i,s]$ denotes the element of the $i$-th station and $s$-th type of passenger flow (inflow/outflow) of $Y_t$.

## IV. CASE STUDY: SHORT-TERM FORECASTING OF STATION-LEVEL INFLOW AND OUTFLOW OF SHENZHEN METRO NETWORK

### A. Dataset Description

a) Passenger flow dataset

The passenger flow dataset was collected from the Automatic Fare Collection (AFC) system of Shenzhen Metro Cooperation[2] in China, which covers a time span of 63 days from Sep. 10th to Nov. 11th in the year of 2013. By 2013, there are 5 metro lines and 118 stations in Shenzhen. The Shenzhen Metro smart card data fields used in this study are listed in Table III.

TABLE III
LIST OF AFC DATA FIELDS

| Field | Description |
| --- | --- |
| Transaction_id | Identifying a transaction |
| Card_id | Identifying a passenger |
| Line_id | Identifying a metro line |
| Station_name | Name of metro station |
| Transaction_type | Indicate either in or out of station |
| Transaction_timestamp | DateTime Timestamp of transaction |

As shown in Table III, a quadruple record (*Card_id, Station_name, Transaction_timestamp, Transaction_type*) can be employed to describe a smart card transaction, where *Card_id* identifies the cardholder (i.e., passenger), *Station_name* refers to a station, *Transaction_timestamp* indicates the timestamp of transaction, and *Transaction_type* identifies either tap-in or tap-out of the station.

Figure 6(a) is the schematic map[3] of Shenzhen metro in 2013, and Figure 6(b) truly reflects the spatial distribution of those stations with the involvement of the geographical information collected from Google map[4].

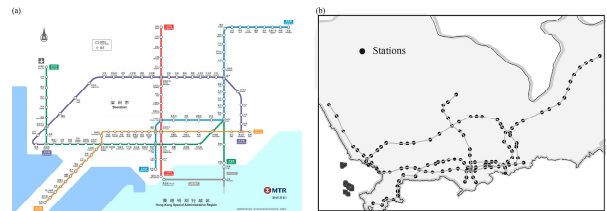

Fig. 6. Shenzhen Metro map of 2013. (a) Schematic map of Shenzhen Metro. (b) Spatial distribution of Shenzhen metro stations.

Concerning the forecasting step, it denotes the granularity of data aggregation in the modeling stage, which is also named analysis interval. Through reviewing the literature about short-term forecasting of subway passenger flow, we found that 15-min was mostly taken as the analysis interval [47-50]. Several studies also used 10-min[51] and 1-hour[52-53] as the analysis interval. On this basis, we determine the analysis interval as 15-min, which is neither too long nor too short for the trade-off between real-time response and stability of the traffic condition. According to the operation of Shenzhen metro system in 2013, the opening hours of all 5 lines are 6:30-23:00. Divide this period by 15-min aggregation interval, we get $\lfloor 16.5h * 60/15 \rfloor = 66$ time slots per day.

---

[2] Shenzhen Metro Cooperation: http://www.szmc.net/
[3] Source: http://toursmaps.com/wp-content/uploads/2017/02/shenzhen_metro_map-1.gif
[4] Source: https://maps.google.com

Three stations are randomly chosen and their inflow and outflow are visualized during each 15-min interval from Nov. 5, 2013 to Nov.11, 2013, as shown in Figure 7(a), (c), and (e). We can see the data oscillated too much, which may make it hard to train the model. Therefore, to reduce the oscillation, we use a one-hour rolling window to move aggregate the 15-min inflow and outflow data. In this way, we can get the recent one-hour inflow/outflow during each 15-min interval. Note that there are 63 time slots per day after moving aggregation since the first three time intervals don't have the aggregated values.

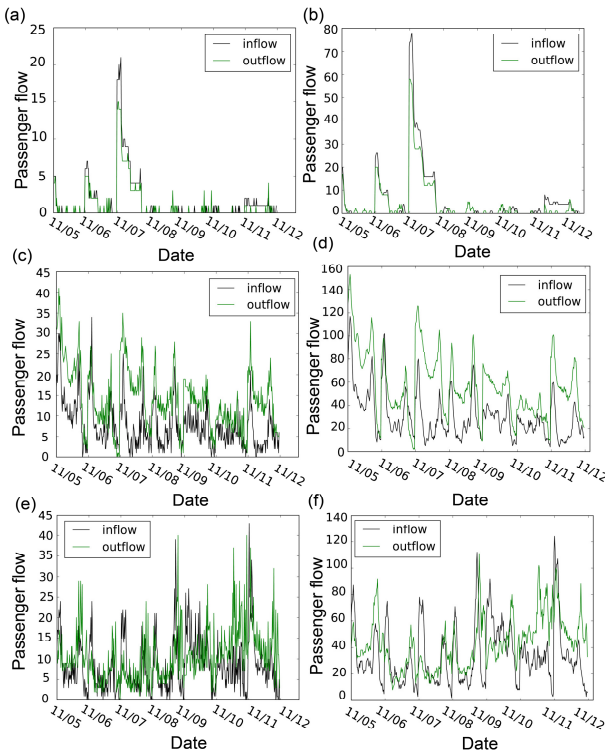

Fig. 7. The raw passenger flow data v.s. the moving aggregated (one-hour sliding window) passenger flow data. (a) The raw passenger flow data of Liyumen station from 2013/11/05 to 2013/11/11. (b) The moving aggregated passenger flow data of Liyumen station from 2013/11/05 to 2013/11/11. (c) The raw passenger flow data of Zhuzilin station from 2013/11/05 to 2013/11/11. (d) The moving aggregated passenger flow data of Zhuzilin station from 2013/11/05 to 2013/11/11. (e) The raw passenger flow data of Shenzhen North Railway station from 2013/11/05 to 2013/11/11. (f) The moving aggregated passenger flow data of Shenzhen North Railway station from 2013/11/05 to 2013/11/11.

As shown in Figure 7(b), (d), and (f), we can see the data are smoother than before, and the model can learn the regulation of data change more easily than before.

We aggregate each station's transaction records (i.e., inflow+outflow) using metro AFC data on October 14. Figure 8 shows the spatial distribution of AFC data transaction records in one day. It presents that the transaction records are most densely distributed at Grand Theatre Station and Laojie Station, closely followed by Huaqiang Road station and Luohu station, and the transaction records of other stations have relatively sparse distribution. From the perspective of the spatial distribution of AFC records in a single day, more latent complex spatial dependencies among stations are pending for being captured.

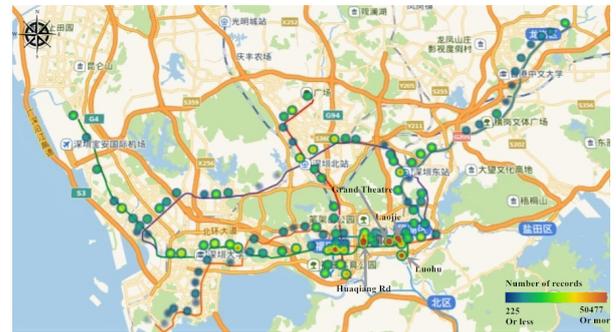

Fig. 8. Spatial distribution of records aggregated at each station during a whole day.

b) POI and network structure dataset

Before collecting the POI-related data, a critical first step is to evaluate the walking distance to metro stations, i.e., the size of a Pedestrian Catchment Areas (PCA), aiming to determine the range of data collection. As the average friendly walking distance is generally assumed to be 500 m in large and middle-sized cities according to Kim et al. (2017) [54], we also define the distance of PCA of each Shenzhen metro station as 500m. In our work, we use a buffer to create circular PCA by 500 m. Based on the buffer with 500m radius determined, POI-related data are collected subsequently. All of the POI data within a PCA were collected from Baidu Map with the assistance of API, and POI data consist of the stations' nearby residence, entertainment, services, business, education, offices. Specifically, the information covers the numbers of residences, restaurants, schools, offices, hospitals, banks, shopping places, bus stations, and hotels within 500m PCA [55].

Network structure data comprise the degree centrality and betweenness centrality of the metro network nodes, days since stations opened, and distance to the city center. Degree centrality and betweenness centrality were calculated according to their definitions and the actual topology of Shenzhen metro network. The information of days since metro lines and stations opened was collected from a website named "UrbanRail"5. As for the distance $Dist_i$ of each station $i$ to the city center, i.e., Shenzhen Municipal People's Government, located in Futian District, we calculate it by the following (10) considering the effect of the radius of the earth:

$$Dist_i = R \times arcos(cos(Lat_0) \times cos(Lat_i) \\ \times cos(Lon_0 - Lon_i) + sin(Lat_i) \\ \times sin(Lat_0)) \times \frac{\pi}{180} \quad (11)$$

The POI-related data and network structure information hypothesized to influence passenger flows of stations are summarized in Table IV.

TABLE IV
SUMMARY OF POI-RELATED DATA AND NETWORK STRUCTURE INFORMATION

| Categories | Explanatory variables | Acronym of variables | Source |
|---|---|---|---|
| POI | No. of residential units | *Residence* | Baidu map |
| | No. of restaurants | *Restaurant* | Baidu map |
| | No. of retailers/shopping | *Shopping* | Baidu map |
| | No. of schools | *School* | Baidu map |
| | No. of offices | *Offices* | Baidu map |
| | No. of banks | *Bank* | Baidu map |
| | No. of hospitals | *Hospital* | Baidu map |
| | No. of hotels | *Hotel* | Baidu map |
| | No. of bus stations | *Bus* | Baidu map |
| | Distance to the city center | *Dis_to_cent* | Calculated |
| Network Structure | Degree centrality | *Degree* | Calculated |
| | Betweenness centrality | *Between* | Calculated |
| | Days since opened | *Days_open* | UrbanRail |

c) Operational information

The information of the metro line's headways of peak hours and off-peak hours is considered as each metro station's operational features. The mean headways[6] of each line in Shenzhen metro are listed in Table V. Note that for the transfer station, the smaller headway is taken as its headway.

TABLE V
THE MEAN HEADWAY OF EACH LINE OF SHENZHEN METRO

| Line | Headway | |
|---|---|---|
| | Peak hours | Off-peak hours |
| L1 | 4 min | 6 min |
| L2 | 8 min | 8 min |
| L3 | 7 min | 8 min |
| L4 | 3 min | 6 min |
| L5 | 6 min | 8 min |

Finally, for the weather factor, the historical weather information of Shenzhen was collected from the website of *Tianqi*[7]. We consider 3 categories of weather variables, including temperature (measured by Celsius degree), weather condition, and wind speed (measured by kilometer per hour). The weather condition variable contains 31 types: such as sunny, clear, fog, rain, etc. We grouped all these types into 8 categories: clear, sunny, fog, cloudy, light rain, rain shower, thunderstorm, and overcast. The weather information we collected takes one value for one hour, so we need to repeat the hourly weather information four times to make it the same length as the other sequences (15-min as the interval).

*B. Data Preprocessing and Settings*

*1) Data Preprocessing (log+diff)*

Time series is different from traditional classification and regression predictive modeling problems. The temporal structure adds order to the observations. When a time series is stationary, it can be easier to model. Statistical modeling methods assume or require the time series to be stationary to be effective. Observations from a non-stationary time series show seasonal effects, trends, and other structures that depend on the time index. Classical time series analysis and forecasting methods are concerned with making non-stationary time series data stationary by identifying and removing trends and removing seasonal effects[49]. Similarly, we also consider removing seasonal and trend effects from the data before feeding them into the deep learning model, to see if it can make model training easier and performing better.

First, log transformation makes data linear and smoother, so we take log transformation to the raw data, as there is zero in raw data, we add 1 to them to make the log calculable.

Besides, differencing makes the data stationary as it removes time series components from the data [56], [57]. The first order differencing takes away only trend, not seasonality. Comparing the results of several orders differencing and observing that there is no clear trend in our data, we take 63th order differencing to remove seasonality, because our data have seasonality at lag=63, i.e., one day has 63 time slots.

In the experiment, we use the passenger flow data after log and differencing transformation from 10/09/2013 to 05/11/2013 (56 days, 3528 time slots) as training data, and the remained 7 days (441 time slots) as testing data.

Moreover, data preparation involves using techniques such as the normalization to rescale input and output variables prior to training a neural network model. Min-Max normalization preserves the shape of the original distribution, and doesn't reduce the importance of outliers. Thus, we use the training data to train the Min-Max scaler and use it to convert the passenger flow data after log and differencing transformation to [0, 1] scale.

After prediction, we denormalize and reverse the differencing and log transformation of the prediction value and use it for evaluation.

*2) Evaluation Metric*

With regards to the evaluation metrics, the most used metrics, i.e., Root Mean Squared Error (RMSE), and MAE are employed (expressed as (11) and (12)). Mean Absolute Percentage Error (MAPE) considers not only the error between the predicted value and the true value but also the ratio between the error and the true value.

However, MAPE has the issue of being infinite or undefined due to zeros in the denominator. Besides, If any true values are very close to zero, the corresponding absolute percentage errors will be extremely high and therefore bias the informativity of the MAPE. The symmetric mean absolute percentage error (sMAPE) was first proposed as a modified MAPE which could be a simple way to fix the issue (expressed as (13)) [58]. It was then used in the M-Competitions (a series of open competitions organized by teams led by forecasting researcher Spyros Makridakis and intended to evaluate and compare the accuracy of different forecasting methods) as an alternative primary measure to MAPE [59].

---

[6] Reference: http://dt.cncn.com/shenzhen/list_time

[7] Website: www.tianqi.com

$$RMSE = \sqrt{\frac{1}{n}\sum_{i=1}^{n}(\hat{y}_i - y_i)^2} \quad (12)$$

$$MAE = \frac{1}{n}\sum_{i=1}^{n}|\hat{y}_i - y_i| \quad (13)$$

$$sMAPE = \frac{1}{n}\sum_{i=1}^{n}\frac{2|\hat{y}_i - y_i|}{(|\hat{y}_i|+|y_i|)} \quad (14)$$

Where $y_i$ and $\hat{y}_i$ are the true value and predicted value, and $n$ is the number of all predicted values.

*3) Model Parameters Determination and Training Settings*

The hyperparameters of MGC-RNN are mainly: learning rate, batch size, the number of epochs at the training stage, the length input sequence, the predicted step the number of graph convolution units, the number of encoder LSTM units, and the number of decoder LSTM units. In the experiment, we manually adjust these hyperparameters. Learning rate decay is a technique for training neural networks. It starts with a large learning rate and then decays it multiple times. An initially large learning rate accelerates training or helps the network escape spurious local minima, and decaying the learning rate helps the network converge to a local minimum and avoid oscillation. So through manual adjustment, we set the learning starts at 0.002, and the decay rate is 0.002 ($lr = lr * \frac{1}{1+decayrate*iterations}$). The batch size is set as 16. The training epoch is set as 50. The input sequence length is 16, and the prediction step is 4 (i.e., 1 hour). We apply 64 graph convolution units for each parallel graph convolutional layer and 200 LSTM units for both encoder and decoder parts. For decoder LSTM, we tried to adopt dropout to avoid overfitting. In LSTM, there are two dropouts applied for different parts of LSTM, wherein dropout rate means the fraction of the units to drop for the linear transformation of the inputs, and recurrent dropout rate means the fraction of the units to drop for the linear transformation of the recurrent state, which drops the connections between the recurrent units[60]. In our experiment, we manually adjust them and set both the dropout and the recurrent dropout rate as 0.2.

The MGC-RNN here is trained by using Adam optimizer [61].

### C. Results and Discussion

*1) Visualization of Weights of Adjacency Matrices of Multigraphs*

First of all, to better understand the rationality of the adjacency matrices of multigraphs calculated by the sparse-based reconstruction model and Gaussian kernel function, we visualize the weights of adjacency matrices $M^1, M^2, M^3, M^4$, and $M_t^5$ of Shenzhen metro in two ways, i.e., map-based correlation, and heatmap. The following Figures 9 and 10 show the visualizations of the four static adjacency matrices

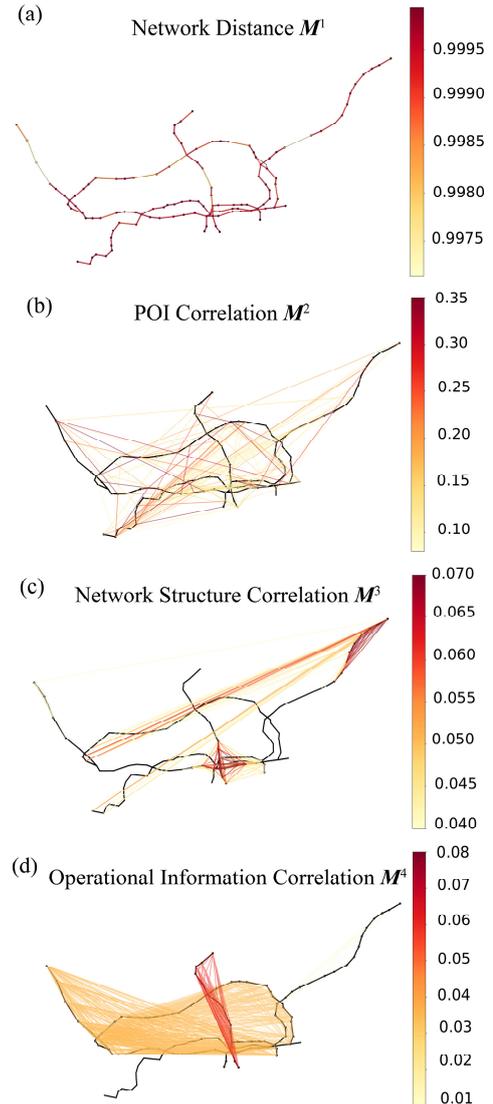

Fig. 9. Map-based visualization of weights of four static adjacency matrices. (a) Weights of network distance adjacency matrix $M^1$. (b) Weights of POI correlation adjacency matrix $M^2$. (c) Weights of network structure correlation adjacency matrix $M^3$. (d) Weights of operational information correlation adjacency matrix $M^4$.

$M^1, M^2, M^3, M^4$ of Shenzhen metro in the form of map-based correlation, and heatmap, respectively. The brighter the color of Figure 9 and the darker the color of Figure 10, the higher the weights. In Figures 9 and 10, to better visualize the weight of each pair of correlated stations, we transform the true values of weights between all station pairs by the log. Besides, the axis labels of Figure 10 consist of both station name and station identifier. For the sake of convenience in representing and understanding, we use alphanumeric code instead of Chinese to denote each station name8. The order of stations is arranged by lines, and the transfer stations are placed at last. Figures 9(a) and 10(a) show the network distance adjacency weights, we can see the nearer the two stations, the weights are higher. For POI correlation (as shown in Figures 9(b) and 10(b)), the weights can be interpreted by combining the land use information of Shenzhen [45]. For network structure correlation (as shown in Figures 9(c) and 10(c)), we can see the central stations and corner stations have higher weights, which is reflected in Figure 10(c) as the blocked distribution because they have similar network structures. For operational information correlation, we can see the higher weights are on the edges between stations in line 4 (as shown in the darkest block of Figure 10(d)) because line 4 is the busies line with the shortest headways in Shenzhen metro.

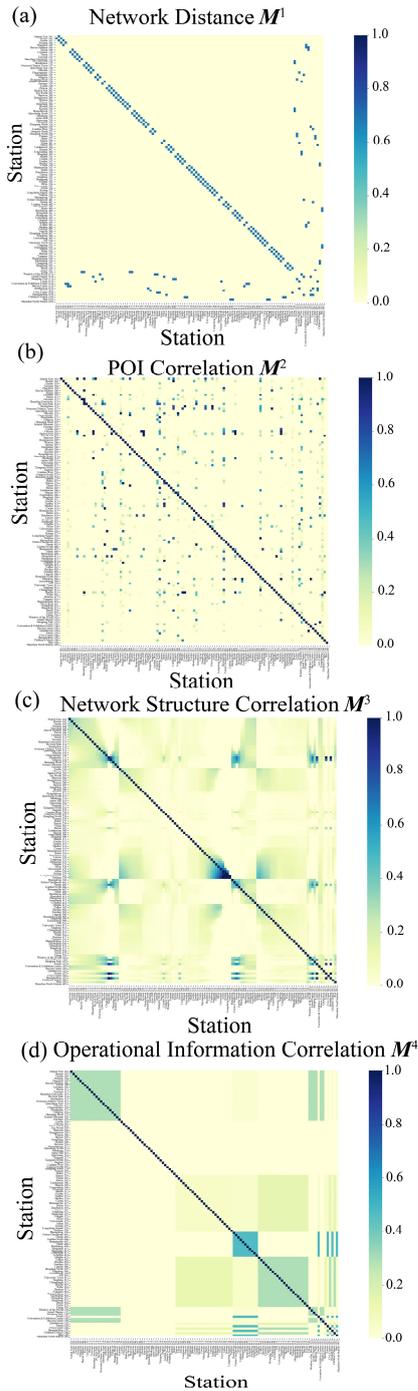

Fig. 10. Heatmap of weights of four static adjacency matrices. (a) Weights of network distance adjacency matrix. (b)Weights of POI correlation adjacency matrix. (c) Weights of network structure correlation adjacency matrix. (d) Weights of operational information correlation adjacency matrix.

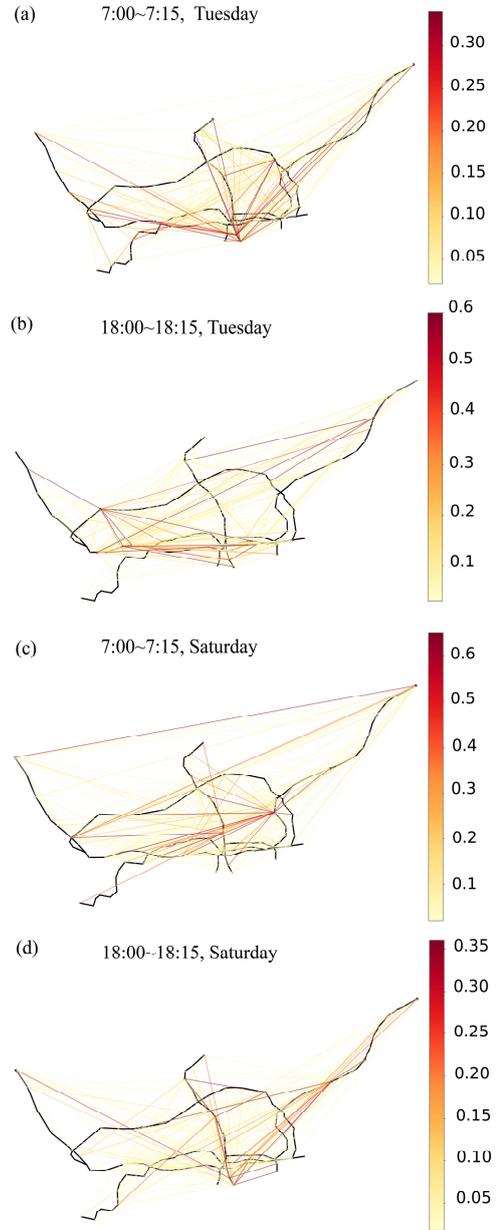

Fig. 11. Map-base visualization of weights of dynamic adjacency matrix during different periods. (a) Weights of recent flow correlation adjacency matrix during 7:00~7:15 on 2013/09/10 (Tuesday). (b) Weights of recent flow correlation adjacency matrix during 18:00~18:15 on 2013/09/10 (Tuesday). (c) Weights of recent flow correlation adjacency matrix during 7:00~7:15 on 2013/09/14 (Saturday). (d) Weights of recent flow correlation adjacency matrix during 18:00~18:15 on 2013/09/14 (Saturday).

---

[8] Here, we define identifiers for station names according to the following rules: (1) non-transfer stations consist of 3 digits, where the first digit denotes the line number, and the rest 2 digits denote the sequential number of station; (2) transfer stations start with character t followed by 3 digits, where the first 2 digits denote the intersection of two lines, and the last digit means the sequential number of intersections between those two lines. For example, "402" represents the 2nd station of Line 4, and "t131" represents the transfer station that is the first intersection of lines 1 and 3. In this way, all 118 stations can be encoded by such identifiers containing the line and station information literally.

With regards to the dynamic adjacency matrix $\boldsymbol{M}_t^5$, the recent flow correlation weights will change over time. We select the morning peak (7:00~7:15) and evening peak (18:00~18:15) of weekday (2013/09/10 (Tuesday)) and weekend (2013/09/14 (Saturday)) to visualize the weights (as shown in Figures 111 and 12). According to Figures 11 and 12, we can observe that some edges with higher weights are changing over time, while some edges are always with higher weights.

To better understand the correlation between stations with higher weights, we visualize the inflow and outflow of two stations with always higher weights in the recent flow correlation adjacency matrix (as shown in Figure 13). Figure 13 (a), Figure 13(b), and Figure 13(c) show the inflow and outflow of Shuanglong v.s. Airport Eastern, Laojie v.s. Guomao, and Grand Theatre v.s. Guomao, respectively. We can see each pair of two stations present quite similar patterns of their inflow and outflow. In this way, we are convinced that the extracted recent flow correlation adjacency matrix is significant to represent the inter-station correlation in terms of passenger flow patterns.

*2) Performance Analysis on Multi-graphs*

First of all, the performance of the models with some of different numbers of alternative combinations of graphs as input was compared (as shown in Table VI, the results of the models with all possible combinations of graphs are shown in Table A1 in Appendix). In addition, in order to see the effect of multigraphs on extracting multi-aspects of features, we also construct a model with only graph, i.e., the network distance graph representing the original topology of the transit network. This model was denoted by GC-RNN, which was also compared with MGC-RNN.

According to Table VI, it is found that the best result is obtained by the MGC-RNN with 3 graphs including $\boldsymbol{M}^2$ POI correlation graph, $\boldsymbol{M}^3$ network structure correlation graph, and $\boldsymbol{M}_t^5$ recent flow correlation graph as the input, which performs much better than the model with 2 graphs without network structure (comparing row 5 with row 11 of Table VI), indicating network structure is very important for the forecasting. Moreover, the model with 4 graphs (rows 2 and 3 of Table VI) or all 5 graphs (row 1 of Table VI) doesn't perform the best, indicating that the model doesn't perform better with more features included. Comparing with row 2 and row 3 in Table V, we may conclude that $\boldsymbol{M}^2$ POI is more important than $\boldsymbol{M}^4$ operational information in MGC-RNN for forecasting passenger flow. Moreover, in order to see if the exogenous factors including day-of-week, holiday information, hour-of-day, and weather information can help improve the performance of passenger flow forecasting, we compare the best MGC-RNN models including 3 graphs ($M^2, M^3, M_t^5$) with different combinations of the exogenous factors and those without exogenous factors (row 5 to row 10). The results show that just including hour_of_day can improve the forecasting accuracy, and the other exogenous factors including holiday, day_of_week, and weather information cannot help to improve the forecasting accuracy but make the performance much worse than before, indicating that day-of-week, holiday information, and weather cannot contribute to the short-term (15-min) forecasting of Shenzhen metro passenger flow. In addition, we also compare MGC-RNN with GC-RNN with $M^1$ and find that using multiple parallel graphs to extract different types of inter-station correlations has a better performance than using one single graph to extract all of the aspects of inter-station correlations excepted for the network distance graph.

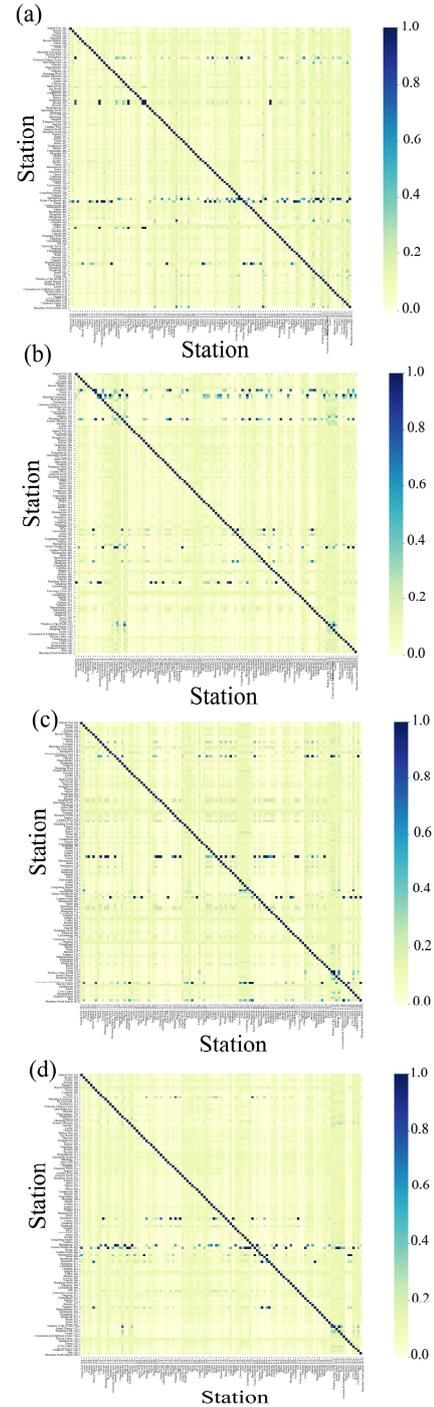

Fig. 12. Heatmaps of weights of dynamic adjacency matrix during different periods. (a) Weights of recent flow correlation adjacency matrix during 7:00~7:15 on 2013/09/10 (Tuesday). (b) Weights of recent flow correlation adjacency matrix during 18:00~18:15 on 2013/09/10 (Tuesday). (c) Weights of recent flow correlation adjacency matrix during 7:00~7:15 on 2013/09/14 (Saturday). (d) Weights of recent flow correlation adjacency matrix during 18:00~18:15 on 2013/09/14 (Saturday).

TABLE VI
THE FORECASTING RESULTS OF MGC-RNN WITH DIFFERENT ALTERNATIVE COMBINATIONS OF GRAPHS AND EXOGENOUS FACTORS AS INPUT)

| | | RMSE | MAE | sMAPE |
|---|---|---|---|---|
| 1 | MGC-RNN (5graphs: $M^1, M^2, M^3, M^4, M_t^5$) | 2.0860 | 1.1911 | 0.0775 |
| 2 | MGC-RNN (4 graphs: $M^1, M^3, M^4, M_t^5$) | 2.3839 | 1.4425 | 0.1000 |
| 3 | MGC-RNN (4 graphs: $M^1, M^2, M^3, M_t^5$) | 2.9042 | 1.5412 | 0.1036 |
| 4 | MGC-RNN (3 graphs: $M^1, M^3, M_t^5$) | 2.1188 | 1.1897 | 0.0795 |
| **5** | **MGC-RNN (3 graphs: $M^2, M^3, M_t^5$)** | **1.8120** | **1.0491** | **0.0749** |
| 6 | MGC-RNN (3 graphs: $M^2, M^3, M_t^5$)_ with $d_t + h_t$ | 8.8791 | 4.5829 | 0.2174 |
| 7 | MGC-RNN (3 graphs: $M^2, M^3, M_t^5$)_ with $o_t + A_t$ | 4.5809 | 2.6095 | 0.1449 |
| 8 | MGC-RNN (3 graphs: $M^2, M^3, M_t^5$)_ with $o_t$ | 1.6425 | 1.0064 | 0.0606 |
| 9 | MGC-RNN (3 graphs: $M^2, M^3, M_t^5$)_ with $A_t$ | 5.0148 | 2.7550 | 0.1475 |
| 10 | MGC-RNN (3 graphs: $M^2, M^3, M_t^5$)_ with $d_t + h_t + o_t + A_t$ | 13.3137 | 7.0529 | 0.2975 |
| 11 | MGC-RNN (2 graphs: $M^2, M_t^5$) | 10.2440 | 5.3959 | 0.2733 |
| 12 | GC-RNN(1 graphs: $M^1$) | 10.6394 | 5.6819 | 0.2765 |

*3) Comparison with Other Forecasting Methods*

We compare the performance of the MGC-RNN model with the following baseline methods.

• **History Average (HA)**: which uses the average passenger inflow and outflow during the previous seasons as the prediction to model the passenger flow as a seasonal process. In this experiment, we take 63 as the period, i.e., 1 day. For example, the average inflow during 7:00~7:15 at station $i$ is estimated by the mean of all historical inflow on all historical 7:00~7:15 of previous days at station $i$. Through compare 63-order HA with our MGC-RNN with 63-order differencing, we can know if the differences trained by our model are more accurate than directly using historical average data.

• **ARIMA**: which is trained on each station's inflow and outflow respectively. The parameters in ARIMA are determined based on Bayesian Information Criterion (BIC)

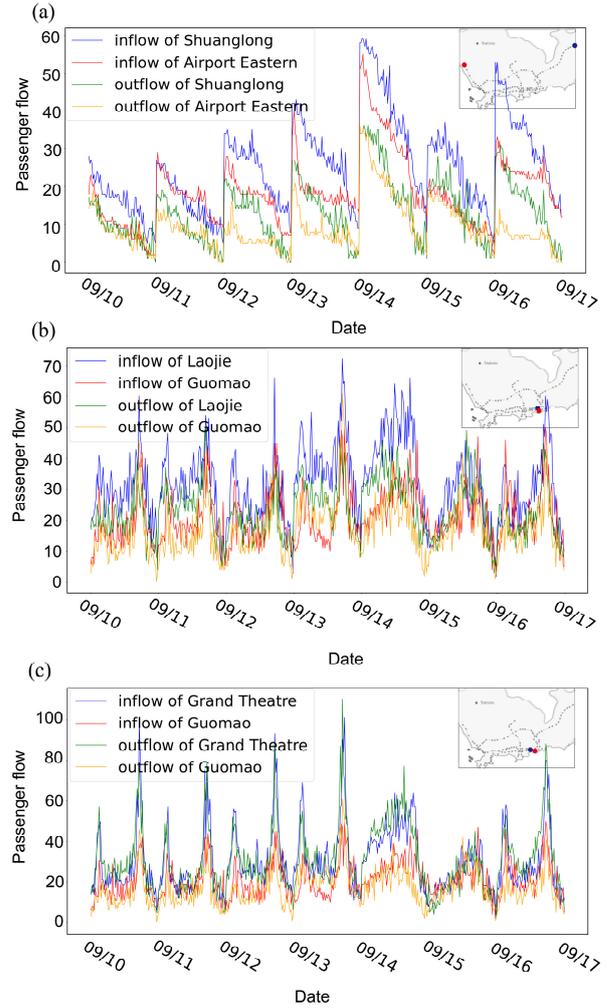

Fig. 13. The inflow and outflow of stations with always higher weights during 2013/09/10~2013/09/16 in the recent flow correlation matrices. (a) Inflow and outflow of Shuanlong v.s. Airport Eastern during 2013/09/10 ~ 2013/09/16. (b) Inflow and outflow of Laojie v.s. Guomao during 2013/09/10 ~ 2013/09/16. (c) Inflow and outflow of Grand Theatre v.s. Guomao during 2013/09/10 ~ 2013/09/16.

[62], Specifically, the parameters are set as $(p, d, q) = (2,0,0)$. To achieve multi-step ahead forecasting, the recursive multi-step forecast strategy is adopted (which uses a one-step model multiple times where the prediction for the prior time step as an input for predicting the following time step). Besides, since ARIMA adopts a dynamic training strategy but not the static training as the regression-based model does, so the training set continually changes as the fixed-length training window slides. Therefore, to try to ensure the fairness of comparison, we have the following two schemes of training:

a) Train on 3528 steps' data, and rolling forecast on 4 steps' data.
b) Train on 16 steps' data, and rolling forecast on 4 steps' data.

• **Vector Auto-Regressive (VAR):** The parameter of VAR, i.e., the lag order $p$ was determined based on the Akaike

Information Criterion (AIC) [63], and $p$ was set as 16. VAR is also trained on 3528 steps' data, and rolling forecast on 4 steps' data. The recursive multi-step forecast strategy is adopted to achieve multi-step forecasting.

- **Least absolute shrinkage and selection operator (LASSO):** which takes 3528 steps' inflow and outflow data as the training set to fit the linear regression model with $\ell 1$ regularization, and takes the remaining data as the testing set. We achieve multivariate forecasting by applying multi-output regression. The recursive multi-step forecast strategy is also adopted to achieve multi-step forecasting.
- **LSTM_encoder-decoder:** to see the effects of multi-graph convolutions in MGC-RNN, LSTM_encoder-decoder is compared with MGC-RNN. We also train it on 3528 steps' inflow and outflow data, and test on the remaining data. The hyperparameters of LSTM_encoder-decoder are: batch_size=16, epochs=50, lstm_seq_len=16, predict_step=4, No. of encoder LSTM units =64, No. of decoder LSTM units= 64, lr=0.001.
- **CNN-LSTM_encoder-decoder:** A popular approach has been to combine CNNs with LSTMs, where the CNN is as an encoder to learn features from sub-sequences of input data which are provided as time steps to an LSTM. This architecture is called a CNN-LSTM. Here, we employed CNN-LSTM architecture to play the encoder and decoder roles, respectively, so we called it as CNN-LSTM_encoder-decoder. We also train it on 3528 steps' inflow and outflow data, and test on the remaining data. The hyperparameters of CNNLSTM_encoder-decoder are: batch_size=16, epochs=50, lstm_seq_len=16, predict_step=4, No. of Conv1D layer=2, No. of Conv1D units in each layer=64, No. of decoder LSTM units= 200, lr=0.001.
- **ConvLSTM_encoder-decoder:** Unlike the CNN-LSTM that is interpreting the output from CNN models, the ConvLSTM is using convolutions directly as part of reading input into the LSTM units themselves. Here, we took a 2D ConvLSTM connecting the Flatten layer as the encoder and LSTM as the decoder. We called it as ConvLSTM_encoder-decoder. We also train it on 3528 steps' inflow and outflow data, and test on the remaining data. The hyperparameters of ConvLSTM_encoder-decoder are: batch_size=16, epochs=50, lstm_seq_len=16, predict_step=4, No. of ConvLSTM2D layer=1, No. of ConvLSTM2D units in each layer=64, No. of decoder LSTM units = 200, lr=0.001.

In the experiment, the input inflow and outflow data fed into all baselines excepted for HA model, are transformed by log and differencing transformation, as well as Min-Max normalization.

Table VII presents the forecasting performance of the best MGC-RNN model (i.e., with 3 graphs including network distance, network structure, and recent flow correlation graph, without exogenous factors) with other baseline methods on passenger flow forecasting, including HA, ARIMA, VAR. LASSO, LSTM_encoder-decoder, CNNLSTM_encoder-decoder, and ConvLSTM_encoder-decoder model.

According to Table VII, we can see the best result is MGC-RNN, and the second-best is LSTM encoder-decoder, indicating the multi-aspects of inter-station correlations extracted by multiple graph convolutions can improve the prediction accuracy of short-term forecasting of passenger flow. Then we can see ARIMA with the training scheme a) performs much better than the scheme b), indicating the training set size matters a lot for ARIMA model. CNNLSTM_encoder-decoder and ConvLSTM_encoder-decoder don't perform better than LSTM_encoder-decoder, indicating that 1D convolutional layers and ConLSTM architecture cannot have a better performance than LSTM layers as the encoders. LASSO and VAR follow and HA performs not well in this study. Since we take 63 as the period when conducting HA, and we take 63-order differencing before training MGC-RNN, through comparing 63-order HA with our MGC-RNN with 63-order differencing, we can know if the differences trained by MGC-RNN model are more accurate than directly using the historical average data. The comparison between HA and MGC-RNN showed in Table VII proves that the differences trained by MGC-RNN indeed work in the forecasting of passenger flow. Moreover, we can see only ARIMA cannot extract inter-station correlations.

TABLE VII
THE FORECASTING RESULTS OF MGC-RNN MODEL AND OTHER BASELINE METHODS

| | Temporal dependences | Inter-station correlation | RMSE | MAE | sMAPE |
|---|---|---|---|---|---|
| HA | √ | × | 26.0591 | 14.0266 | 0.5718 |
| ARIMA (scheme a) | √ | × | 11.5164 | 6.8897 | 0.3740 |
| ARIMA (scheme b) | √ | × | 64.4251 | 13.5372 | 0.4439 |
| VAR | √ | √ | 15.6633 | 9.7980 | 0.5208 |
| LASSO | √ | √ | 15.2971 | 6.7762 | 0.3725 |
| MGC-RNN (3 graphs: $M^2, M^3, M_t^5$)) | √ | √ | 1.8120 | 1.0491 | 0.0749 |
| LSTM_encoder-decoder | √ | √ | 11.4319 | 6.1651 | 0.2978 |
| CNNLSTM_encoder-decoder | √ | √ | 13.6672 | 7.2184 | 0.3253 |
| ConvLSTM_encoder-decoder | √ | √ | 14.9209 | 8.1140 | 0.3679 |

Also, compared with other related studies on passenger flow forecasting of Shenzhen metro [64]–[68], our model outperforms their models in terms of prediction accuracy, for both single stations and overall network.

**Fairness of comparison**

Furthermore, the fairness of comparison among forecasting models is discussed. There are two kinds of training schemes of forecasting models: dynamic training scheme and static training scheme.

For dynamic training models using the sliding window strategy, such as the ARIMA and VAR models, the training set continually changes as the fixed-length training window slides.

For static training models, they are often regression-based models, such as linear regression, LASSO, ridge regression, and deep learning-based models, they have fixed training and testing sets. The trained model is obtained during a "one-time" training stage via several epochs and then used to predict all of the cases in the testing set.

The biggest weakness of static models compared with dynamic models is that if some stations' passenger flows present the opposite scenarios in the training and testing stages, the static models may have poor forecasting results for such certain stations, for which the trained static models do not accurately predict the testing data.

On this basis, we may compare these methods more objectively from the perspective of the training mechanism of these methods.

*4) Discussion on the Performance of Each Prediction Step*

Since we conduct multistep forecasting, we can obtain the 4 steps of predicted results at one time. Here, we randomly chose several stations in each prediction step to visualize the prediction results. Besides, according to the number of passenger flows, we can categorize the stations into busy and non-busy stations groups (Through analyzing the empirical cumulative distribution function (CDF), we took 150 as the boundary, i.e., stations with a maximum passenger flow of less IEEE TRANSACTIONS ON INTELLIGENT TRANSPORTATION SYSTEMS 18than 150 are classified as non-busy stations, and vice versa). Therefore, we can observe the predicted results of randomly chosen stations at each predicted step. Figure 14 shows four busy stations' predicted results at one prediction step by MGC-RNN with 3 graphs including network distance, network structure, and recent flow correlation. Figure 15 shows four non-busy stations' predicted results at one prediction step by MGC-RNN with 3 graphs including network distance, network structure, and recent flow correlation.

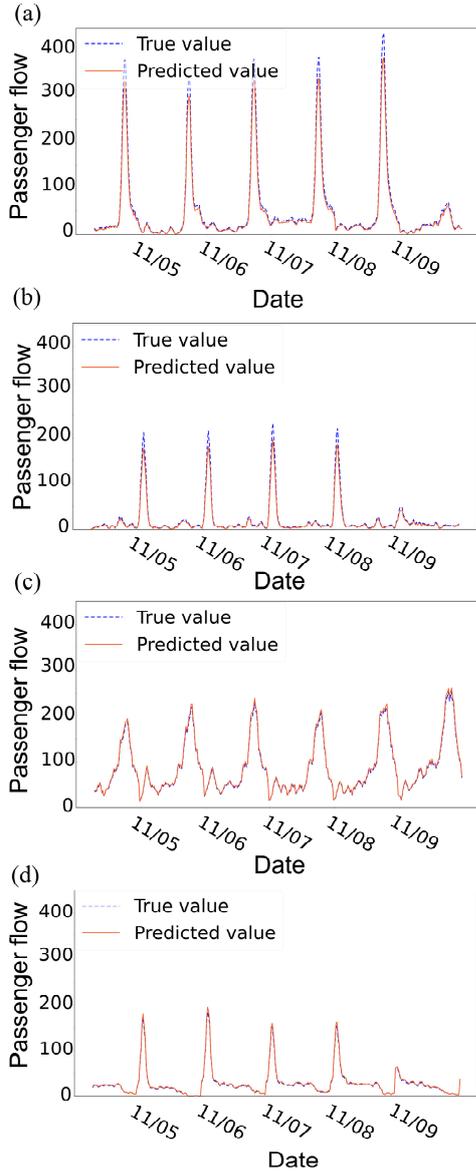

Fig. 14. The predicted results for four randomly chosen busy stations at one predicted step. (a) Predicted outflow v.s. the true value of Shenzhen University station at the first forecasting step. (b) Predicted inflow v.s. the true value of Shenzhen University station at the second forecasting step. (c) Predicted outflow v.s. the true value of Futian Checkpoint station at the third forecasting step. (d) . Predicted inflow v.s. the true value of Futian station at the fourth forecasting step.

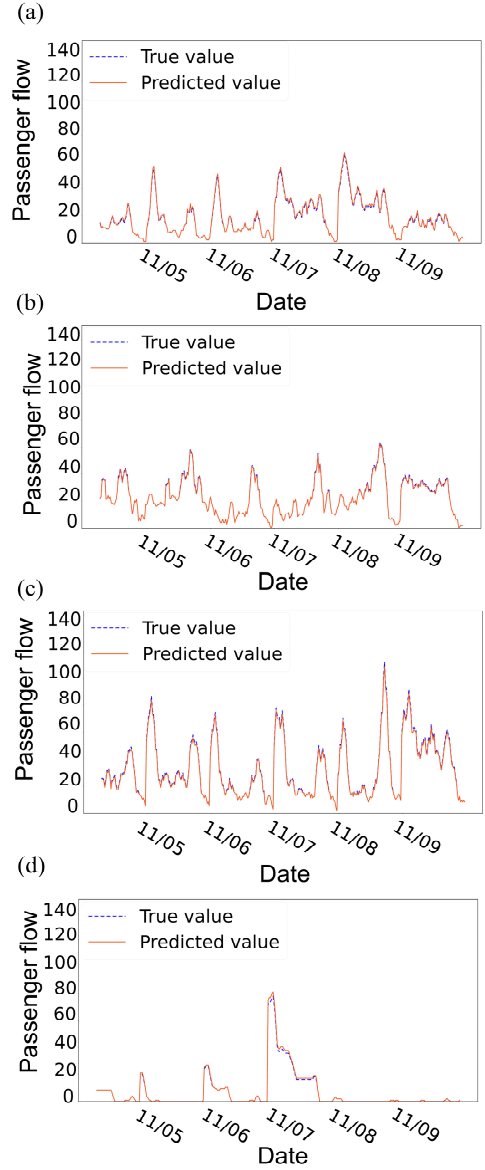

Fig. 15. The predicted results for four randomly chosen non-busy stations at one predicted step. (a) Predicted outflow v.s. the true value of Lianhuacun station at the first forecasting step. (b) Predicted outflow v.s. the true value of Liuxiandong station at the second forecasting step. (c) Predicted inflow v.s. the true value of Shenzhen Northern Railway station at the third forecasting step. (d) . Predicted inflow v.s. the true value of Liyumen station at the fourth forecasting step.

Through these two figures, we can see our model can capture the passenger flow patterns accurately no matter of busy stations or non-busy stations. Even for the sudden extreme peak such as that at Liyumen station, we can accurately detect it in advance.

Besides, to observe each-step prediction errors, we choose one representative of models using recursive multi-step forecast strategy to conduct multi-step forecasts, i.e., ARIMA, and compare it with MGC-RNN model with 3 graphs including network distance, network structure, and recent flow correlation graph, which uses Seq2seq architecture to achieve multi-step forecasts. The results of MGC-RNN and ARIMA with the training scheme a) at each predicted step are shown in Table VIII.

TABLE VIII
THE FORECASTING RESULTS OF MGC-RNN WITH 3 GRAPHS AND ARIMA WITH THE TRAINING SCHEME A) AT EACH PREDICTION STEP

| | RMSE | MAE | sMAPE |
|---|---|---|---|
| MGC-RNN ($M^1, M^3, M_t^5$) ($p^* = 1$) | 2.1188 | 1.1897 | 0.0795 |
| MGC-RNN ($M^1, M^3, M_t^5$) ($p = 2$) | 2.1182 | 1.1889 | 0.0795 |
| MGC-RNN ($M^1, M^3, M_t^5$) ($p = 3$) | 2.1176 | 1.1880 | 0.0796 |
| MGC-RNN ($M^1, M^3, M_t^5$) ($p = 4$) | 2.1177 | 1.1881 | 0.0796 |
| ARIMA (scheme a) ($p = 1$) | 5.4097 | 3.2709 | 0.2488 |
| ARIMA (scheme a) ($p = 2$) | 9.5009 | 5.8837 | 0.3438 |
| ARIMA (scheme a) ($p = 3$) | 12.7339 | 8.1675 | 0.4214 |
| ARIMA (scheme a) ($p = 4$) | 15.6758 | 10.2366 | 0.4822 |

According to Table VIII, we can see the prediction errors of ARIMA accumulate as the forecasting step increases, but errors of MGC-RNN don't propagate. The error propagation is not presented by Seq2Seq architecture in our four-step forecasting task (not really further-step), indicating the strength of Seq2Seq architecture compared with the traditional recursive multi-step forecast strategy in multi-step forecasting.

## V. CONCLUSION

In this paper, we propose a novel deep learning approach, named Multi-Graph Convolutional-Recurrent Neural Network (MGC-RNN), to consider spatiotemporal dependencies and the complex inter-station correlations measured by static and dynamic factors simultaneously in the short-term forecasting of passenger flow. Specifically, we generate multiple graphs (including static and dynamic) to represent the inter-station correlations driven by different factors, respectively. Then we apply multiple GCNs to extract each graph's correlation information and then weighted-fuse all the extracted information. With regards to the temporal dependencies, one of Seq2Seq architecture, LSTM_encoder-decoder was employed to extract temporal dependencies and achieve the multi-step forecasting. Furthermore, MGC-RNN can also include exogenous factors such as national public holidays, and the information of day-of-week by processing through embedding and F-C layers.

MGC-RNN is validated on a real-world dataset, the Shenzhen metro smart card data collected from AFC system. Our model is also compared with several benchmark algorithms, including the traditional time series methods such as HA, ARIMA, VAR, machine learning methods, LASSO, and the popular used forecasting neural network, LSTM_encoder-decoder. The results show that the MGC-RNN outperforms the benchmark algorithms in the measurements of RMSE, MAE, and sMAPE significantly, indicating that the proposed approach performs better at capturing the complex inter-station correlations and temporal dependencies for passenger inflow and outflow forecasting of all stations together. In the experiment, we found that the inter-station driven by network distance, network structure, and recent flow patterns are significant factors for passenger flow forecasting. Exogenous factors including day-of-week and holiday information cannot help to improve the prediction accuracy. Besides, the architecture of LSTM-encoder-decoder can not only capture the temporal dependencies well but also make predictions at each predicted step accurately. The error propagation doesn't present in our model for the four-step forecasting, indicating the strength of Seq2Seq architecture compared with the traditional recursive multi-step forecast strategy in multi-step forecasting.

Above all, our proposed deep learning framework has presented the capability of short-term passenger flow forecasting multi-step ahead with high accuracy. Moreover, the proposed model structure is expected to be flexible enough to handle several similar spatiotemporal forecast tasks, e.g., traffic state forecasting in other transportation systems, infectious disease forecasting, and weather conditions forecasting, etc. In practical scenarios, the proposed framework could provide multiple views of passenger flow dynamics for fine prediction. Specifically, by incorporating various types of inter-station correlations, temporal dependencies, and exogenous factors, the framework exhibits a possibility for multi-source heterogeneous data fusion in a big data environment. With the accurate passenger inflow and outflow in the immediate future forecasted through our proposed framework, it is possible for transit operators to be aware of an emergency (an influx of passengers) and implement emergency preparedness plans in advance, optimize service schedules, and enhance station passenger crowd regulation planning. For passengers, an accurate short-term forecast of passenger flow information can help them to adjust their travel paths, modes, and departure times flexibly and rationally. Thus, systematic objectives related to safety, high efficiency, and service quality can be achieved. In the future, we will try to find some more exogenous factors in the model to help to capture the abnormal patterns of passenger flow. More work will be done on the passenger flow forecasting under uncertainties.

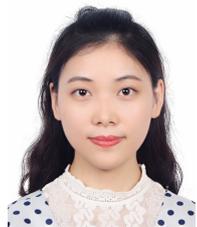

**Yuxin He** earned her B.Eng. and M.Sc. degrees in transportation engineering from Central South University, Changsha, China, in 2014 and 2017, respectively. She received her Ph.D. degree from City University of Hong Kong in data science in 2020. She is currently an Assistant Professor with the College of Urban Transportation and Logistics, Shenzhen Technology University. Her research interests include data mining, traffic flow analysis, modelling and forecasting, and network analysis.

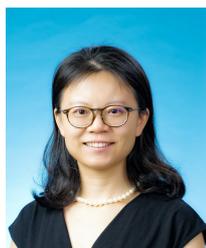

**Lishuai Li** received her B.Eng. degree in aircraft design and engineering from Fudan University, and her M.Sc. and Ph.D. degrees in air transportation systems from the Department of Aeronautics and Astronautics, Massachusetts Institute of Technology (MIT). She is currently an Assistant Professor, Section of Air Transport Operations, Faculty of Aerospace Engineering, Delft University of Technology, and affiliated with the School of Data Science, City University of Hong Kong. Her research interests are in the development of analytical methods for the design, management, and operation of transportation systems, focusing on air transport.

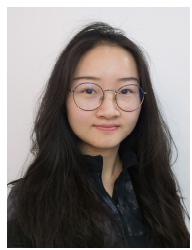

**Xinting Zhu** received her B.Eng. degree in aircraft airworthiness engineering from Beihang University in 2018. She is currently pursuing her Ph.D. degree with the Department of Systems Engineering and Engineering Management, City University of Hong Kong. Her research interests are data analytics, machine learning, deep learning, and their application to air transportation operation and management.

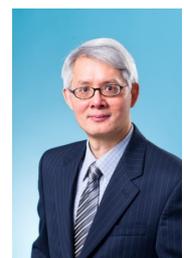

**Kwok Leung Tsui** is Professor at Department of Industrial and Systems Engineering, Virginia Polytechnic Institute and State University. Prior to joining Virginia Polytechnic Institute and State University, Professor Tsui was Chair Professor of Industrial Engineering with School of Data Science, City University of Hong Kong, the founder and Director of Center for Systems Informatics Engineering, and Professor at the School of Industrial and Systems Engineering at the Georgia Institute of Technology. Professor Tsui was a recipient of the National Science Foundation Young Investigator Award. He is Fellow of the American Statistical Association, American Society for Quality, and International Society of Engineering Asset Management. Professor Tsui's current research interests include data mining, surveillance in healthcare and public health, prognostics and systems health management, calibration and validation of computer models, process control and monitoring, and robust design and Taguchi methods.

## VI. Appendix

TABLE A1
THE FORECASTING RESULTS OF MGC-RNN WITH ALL ALTERNATIVE COMBINATIONS OF GRAPHS AS INPUT

| # | Model | RMSE | MAE | sMAPE |
|---|---|---|---|---|
| 1 | MGC-RNN (5 graphs: $M^1, M^2, M^3, M^4, M_t^5$) | 2.0860 | 1.1911 | 0.0775 |
| 2 | MGC-RNN (4 graphs: $M^1, M^3, M^4, M_t^5$) | 2.3839 | 1.4425 | 0.1000 |
| 3 | MGC-RNN (4 graphs: $M^1, M^2, M^3, M_t^5$) | 2.9042 | 1.5412 | 0.1036 |
| 4 | MGC-RNN (4 graphs: $M^1, M^2, M^4, M_t^5$) | 2.7628 | 1.5475 | 0.1044 |
| 5 | MGC-RNN (4 graphs: $M^2, M^3, M^4, M_t^5$) | 2.2292 | 1.2842 | 0.0857 |
| 6 | MGC-RNN (4 graphs: $M^1, M^2, M^3, M^4$) | 2.6375 | 1.4712 | 0.1082 |
| 7 | MGC-RNN (3 graphs: $M^1, M^2, M_t^5$) | 1.9244 | 1.1074 | 0.0770 |
| 8 | MGC-RNN (3 graphs: $M^1, M^3, M_t^5$) | 2.1188 | 1.1897 | 0.0795 |
| 9 | MGC-RNN (3 graphs: $M^1, M^4, M_t^5$) | 3.4844 | 1.8968 | 0.1172 |
| 10 | MGC-RNN (3 graphs: $M^2, M^3, M_t^5$) | 1.8120 | 1.0491 | 0.0749 |
| 11 | MGC-RNN (3 graphs: $M^2, M^4, M_t^5$) | 2.1859 | 1.2526 | 0.1065 |
| 12 | MGC-RNN (3 graphs: $M^3, M^4, M_t^5$) | 2.5887 | 1.4922 | 0.0949 |
| 13 | MGC-RNN (3 graphs: $M^1, M^2, M^3$) | 4.2586 | 2.0843 | 0.1026 |
| 14 | MGC-RNN (3 graphs: $M^1, M^2, M^4$) | 2.5589 | 1.4334 | 0.0846 |
| 15 | MGC-RNN (3 graphs: $M^2, M^3, M^4$) | 2.1518 | 1.1867 | 0.1073 |
| 16 | MGC-RNN (3 graphs: $M^1, M^3, M^4$) | 2.4570 | 1.5115 | 0.0832 |
| 17 | MGC-RNN (2 graphs: $M^1, M_t^5$) | 9.8736 | 5.1729 | 0.2563 |
| 18 | MGC-RNN (2 graphs: $M^2, M_t^5$) | 10.2440 | 5.3959 | 0.2733 |
| 19 | MGC-RNN (2 graphs: $M^3, M_t^5$) | 9.9212 | 5.2014 | 0.2549 |
| 20 | MGC-RNN (2 graphs: $M^4, M_t^5$) | 10.0626 | 5.2699 | 0.2595 |
| 21 | MGC-RNN (2 graphs: $M^1, M^2$) | 9.8816 | 5.1361 | 0.2608 |
| 22 | MGC-RNN (2 graphs: $M^1, M^3$) | 10.9160 | 5.6441 | 0.2818 |
| 23 | MGC-RNN (2 graphs: $M^1, M^4$) | 9.6442 | 5.1670 | 0.2629 |
| 24 | MGC-RNN (2 graphs: $M^2, M^3$) | 10.2196 | 5.4642 | 0.2676 |
| 25 | MGC-RNN (2 graphs: $M^2, M^4$) | 9.9739 | 5.3045 | 0.2670 |
| 26 | MGC-RNN (2 graphs: $M^3, M^4$) | 10.0985 | 5.3235 | 0.2622 |
| 27 | GC-RNN (1 graphs: $M^1$) | 10.6394 | 5.6819 | 0.2765 |
| 28 | GC-RNN (1 graphs: $M^2$) | 10.3505 | 5.4339 | 0.2702 |
| 29 | GC-RNN (1 graphs: $M^3$) | 10.2013 | 5.4231 | 0.2721 |
| 30 | GC-RNN (1 graphs: $M^4$) | 10.1453 | 5.2205 | 0.2557 |
| 31 | GC-RNN (1 graphs: $M_t^5$) | 10.0668 | 5.3790 | 0.2686 |